%% file: neurips_2023.tex
\documentclass{article}

\usepackage{preamble}

\input{text/header}
\begin{document}
\maketitle
\input{text/0-abstract}
\input{text/1-introduction}

\input{text/2-preliminary}
\input{text/3-removal-based-explanation}
\input{text/4-methodology}

\input{text/5-experiments}
\input{text/7-conclusion}

\bibliographystyle{plainnat} 
\bibliography{regex}
\input{text/8-appendix}

\end{document}

%% file: text/header.tex
\title{Efficient GNN Explanation via Learning Removal-based Attribution}

\author{%
Yao Rong$^{1}$, Guanchu Wang$^{2}$, Qizhang Feng$^{3}$, Ninghao Liu$^{4}$, Zirui Liu$^{2}$, \\
\textbf{Enkelejda Kasneci$^{1}$,  Xia Hu$^{2}$} \\
$^1$Technical University of Munich, $^2$Rice University \\ $^3$Texas A\&M University, $^4$University of Georgia \\
\scriptsize \texttt{\{yao.rong,enkelejda.kasneci\}@tum.de}, \texttt{\{guanchu.wang,zl105,xia.hu\}@rice.edu}  \\
\scriptsize  \texttt{qf31@tamu.edu}, \texttt{ninghao.liu@uga.edu}
}

%% file: text/0-abstract.tex



\begin{abstract}
As Graph Neural Networks (GNNs) have been widely used in real-world applications, model explanations are required not only by users but also by legal regulations. However, simultaneously achieving high fidelity and low computational costs in generating explanations has been a challenge for current methods.
In this work, we propose a framework of GNN explanation named~\Algname{}~(\Algnameabbr{}) to address this problem. 
Specifically, we introduce removal-based attribution and demonstrate its substantiated link to interpretability fidelity theoretically and experimentally. The explainer in \Algnameabbr{} learns to generate removal-based attribution which enables providing explanations with high fidelity.
A strategy of subgraph sampling is designed in \Algnameabbr{} to improve the scalability of the training process.
In the deployment, \Algnameabbr{} can efficiently generate the explanation through a feed-forward pass. 
We benchmark our approach with other state-of-the-art GNN explanation methods on six datasets. Results highlight the effectiveness of our framework regarding both efficiency and fidelity. In particular, \Algnameabbr{} is 3.5$\times$ faster and achieves higher fidelity than the state-of-the-art method on the large dataset ogbn-arxiv~(more than 160K nodes and 1M edges), showing its great potential in real-world applications. Our source code is available at \url{https://anonymous.4open.science/r/LARA-10D8/README.md}.
\end{abstract}

%% file: text/1-introduction.tex
\section{Introduction}

Graph Neural Networks (GNNs) achieve great success in many areas, while their black-box nature limits their applications where a trustworthy decision is needed, such as in recommender systems~\cite{wu2022graph}, social networks~\cite{fan2019graph}, and biochemical or medical analysis~\cite{wu2020comprehensive}.
In many scenarios, it is therefore important to explain the behavior of GNNs due to the requirement of both stakeholders and regulations.
For instance, according to the European Union General Data Protection Regulation~\cite{gdpr}, commercial recommender systems should provide users with explanations along with recommendations.


In recent years, various methods have been proposed for GNN explanation~\cite{yuan2022explainability}.
There are two major aspects for evaluating the real-world applicability of an explanation method: \textbf{fidelity}~\cite{yuan2022explainability, liu2022interpretability, zhang2022gstarx,rong2022consistent} and \textbf{efficiency}~\cite{chuang2023efficient, wang2022accelerating, chuang2022cortx, liu2022exact, liu2022rsc}. The fidelity reflects the faithfulness of explanation~\cite{yuan2022explainability}, and the efficiency indicates the latency of a method deployed to real-world scenarios~\cite {chuang2023efficient}.
To generate faithful explanations, existing approaches rely on a searching process of input subgraph that can minimize the variance of GNN output~\cite{liu2022interpretability, feng2021degree}.
However, these pieces of work suffer from the time-consuming process of searching the explanations.
For instance, to obtain an explanation, GNNExplainer~\cite{ying2019gnnexplainer} requires learning a mask of input neighbors; ZORRO~\cite{funke2020hard} and SubgraphX~\cite{yuan2021explainability} utilize on greedy algorithms and Monte Carlo Tree Search to sample important subgraph that can maximize the designed objective function.
Either the learning or searching overhead can significantly slow down the process of explanation generation, which limits their application to real-time scenarios. This is particularly relevant in social network graphs that frequently consist of millions of nodes and edges.

In order to provide efficient GNN explanations, previous work~\cite{luo2020parameterized,lin2021generative, vu2020pgm, chuang2022cortx} attempts to train a deep neural network-based explainer, amortizing the time and resource cost of generating explanations of many samples. However, it is quite challenging to ensure the faithfulness/fidelity of explanations due to the absence of ``ground-truth explanations'' to provide the supervision of faithfulness in the training process. 
Existing work suggests maximizing the mutual information (MI) between the prediction made using learnable explanatory subgraphs and the original graphs, but the MI value fails to be consistent with the interpretation fidelity, especially on large graph datasets where the number of neighbors increases.
Another challenge is to design a scalable training process considering the limited memory capacity of computational units~(GPU), as training the explainer requires huge memory to store the graph structure for aggregation operation. Training on large graphs should be deployed with optimization to avoid exceeding the memory capacity.

\begin{wrapfigure}{R}{0.43\textwidth}
  \scalebox{0.98}{
    \begin{minipage}{.97\linewidth}
        \vspace{-8pt}
        \centering
        \includegraphics[width=1.0\linewidth]{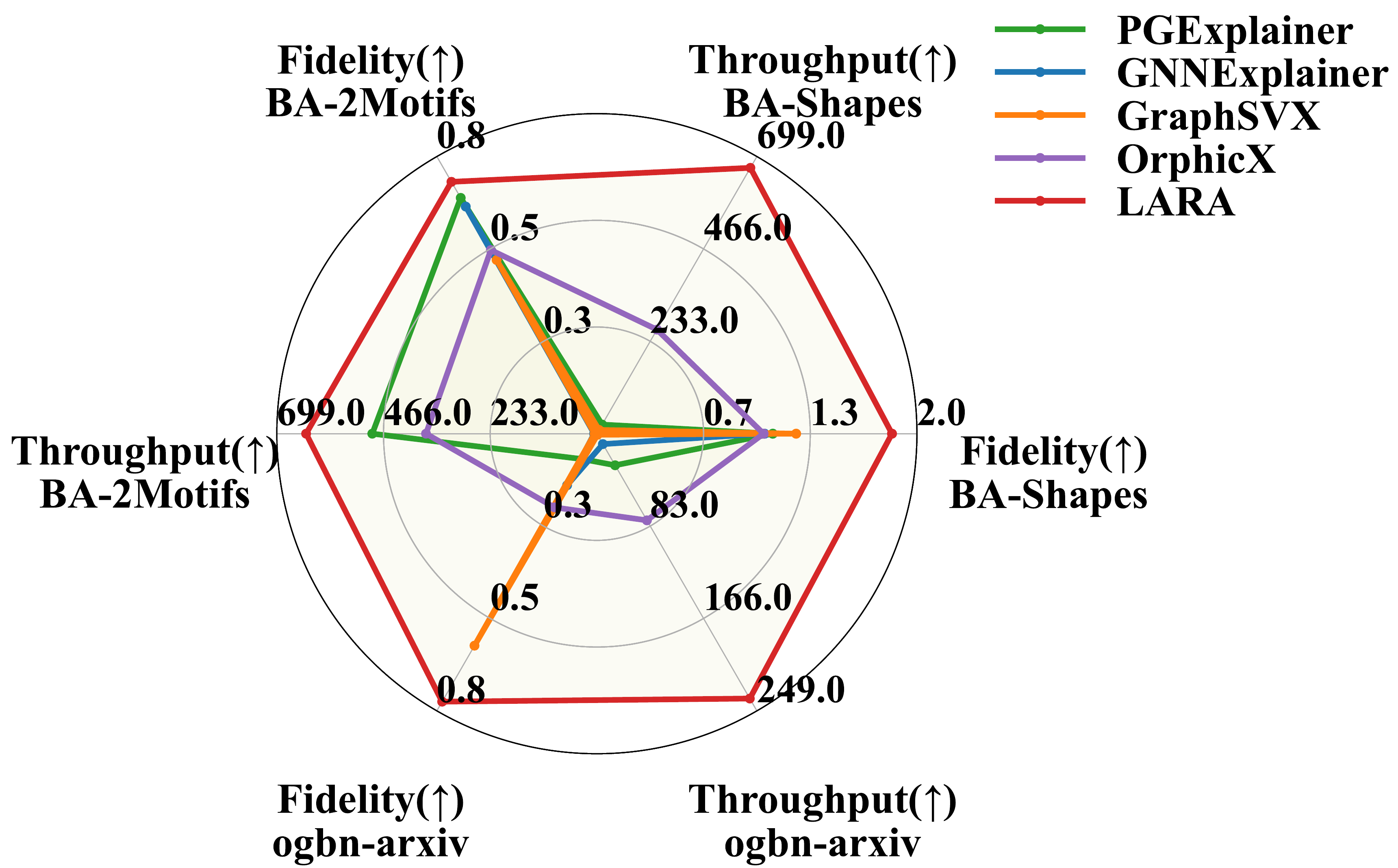}
        \caption{\small Comparison of \Algnameabbr{} with state-of-the-art GNN explanations in fidelity and efficiency on the BA-2Motifs~(graph classification), BA-Shapes~(node classification), and ogbn-arxiv~(node classification). Experimental details in this figure are given in Appendix~\ref{supp:details_radar}.} 
        \label{fig:radar}
        \vspace{-8pt}
    \end{minipage}
    }
  \end{wrapfigure}



To bridge this research gap, we propose a framework \Algname{} (\Algnameabbr{}) towards efficient and faithful GNN explanation. 
Specifically, we first experimentally demonstrate the MI value becomes less correlated with fidelity as the number of neighbors grows. Thus, we propose removal-based attribution for GNN explanation, and demonstrate its contribution to the high fidelity of interpretation from both theoretical and empirical perspectives. Then, we develop an amortized explainer to learn the removal-based attribution, such that it can efficiently generate GNN explanation in a feed-forward process on complex graph structures.
Finally, we design a strategy of subgraph sampling in cooperation with the training of explainer to improve its scalability on large graphs.
Experiments on six benchmark datasets demonstrate the effectiveness of \Algnameabbr{} regarding fidelity, efficiency, and scalability.

The ever-growing real-world social graphs can have more than millions of nodes and edges, poses significant challenges for the efficiency and scalability of GNN explanation approaches.
According to the overall evaluation of fidelity and efficiency in \Cref{fig:radar}, on the ogbn-arxiv dataset with over 160K nodes and 1M edges, \textbf{\Algnameabbr{} achieves a $3.5\times$ speedup as well as higher interpretation fidelity} than the state-of-the-art approach.
This indicates the significant potential of our work for real-world applications.
To summarize, our work makes the following contributions:
\begin{itemize}[leftmargin=10pt, topsep=1pt, itemsep=0pt]
    \item We experimentally study the inconsistency of mutual information with interpretation fidelity as the number of neighbors increases. We propose removal-based attribution for GNN explanation, and demonstrate its contribution to high fidelity from both theoretical and experimental perspectives.

    \item We propose a framework \Algnameabbr{} which deploys an amortized explainer for efficient GNN explanation generation. The learning of the explainer incorporates the proposed removal-based attribution towards fidelity maximization, along with a subgraph sampling strategy for training scalability. 
        

    \item We benchmark \Algnameabbr{} on six datasets including a real-world large graph dataset, in comparison with state-of-the-art methods. The experiment results demonstrate the effectiveness of \Algnameabbr{} in terms of fidelity, efficiency, and scalability.
\end{itemize}

%% file: text/2-preliminary.tex
\section{Preliminaries}
In this section, we introduce the notations for the problem formulation and evaluation metrics that motivate the design of our GNN explanation framework. 

\subsection{Notations}
We consider an arbitrary black-box GNN $f(\cdot)$ as the target model to be explained.
The GNN model $f(\cdot)$ is trained on the graph $\mathcal{G} = (\mathcal{V}, \mathcal{E})$, where $\mathcal{V} =\{v_1, ..., v_N\}$ and $\mathcal{E}$ denote the node set and edge set, respectively; $N = |\mathcal{V}|$ is the number of nodes in the graph.
For any node subset $\mathcal{S} \!\subseteq\! \mathcal{V}$, let $\mathcal{G}_{\mathcal{S}} = \{ \mathcal{S}, \mathcal{E}_{\mathcal{S}} \}$ denote the subgraph which consists of the nodes in $\mathcal{S}$ and the related edges $\mathcal{E}_{\mathcal{S}}$, where $\mathcal{E}_{\mathcal{S}}$ represents the edges connecting the nodes in $\mathcal{S}$.
For each node $v_i \in \mathcal{V}$, let $\mathcal{N}(v_i)$ denote the neighbors of node $v_i$ including itself~(as the zero-hop neighbor). 
Thus, $\mathcal{G}_{\mathcal{N}(v_i)} = \{ \mathcal{N}(v_i), \mathcal{E}_{\mathcal{N}(v_i)} \}$ denotes the subgraph containing the neighbors of $v_i$ and their associated edges;
and $f(\mathcal{G}_{\mathcal{N}(v_i)})$ aggregates the information from its neighbors to make the prediction of node $v_i$.
In this work, we focus on estimating the attribution of each neighbor $v_j \in \mathcal{N}(v_i)$ to the GNN prediction $f(\mathcal{G}_{\mathcal{N}(v_i)})$.

\subsection{Explanation Evaluation Metrics}
\label{sec:eval metrics}
\paragraph{Fidelity.} 
We follow existing work~\cite{yuan2022explainability, zhang2022gstarx, liu2022interpretability} to consider $\mathrm{Fidelity}^+$ and $\mathrm{Fidelity}^-$ as the metrics to evaluate GNN explanation in this work.
Concretely, fidelity reflects the faithfulness of explanations~\cite{tomsett2020sanity,yeh2019fidelity,alvarez2018towards}, which can be widely applied to scenarios where ground truth explanations are not available.
Formally, given the important neighbors $\mathcal{N}^*_{v_i} \subseteq \mathcal{N}(v_i)$ of a node $v_i$ discovered by an explanation method, $\mathcal{G}_{\mathcal{N}^*_{v_i}}$ denotes the subgraph consisting of the important nodes $\mathcal{N}^*_{v_i}$ and the associated edges. Likewise, $\mathcal{G}_{\mathcal{N}(v_i) \setminus \mathcal{N}^*_{v_i}}$ represents the subgraph after removing the important nodes $\mathcal{N}^*_{v_i}$ and their associated edges from $\mathcal{G}_{\mathcal{N}(v_i)}$, i.e., unimportant nodes are preserved in $\mathcal{N}(v_i) \setminus \mathcal{N}^*_{v_i}$.


$\mathrm{Fidelity}^+$ evaluates the explanation method via \textit{removing the important neighbors} and checking the model prediction changes. 
Therefore, $\mathrm{Fidelity}^+$ is defined as:
\begin{equation}
 \uparrow \mathrm{Fidelity}^+  = \frac{1}{N}\sum_{i=1}^N f(\mathcal{G}_{\mathcal{N}(v_i)}) - f( \mathcal{G}_{\mathcal{N} (v_i) \setminus \mathcal{N}^*_{v_i}} ),
  \label{eq:fidelity+}
\end{equation}
where $f(\mathcal{G}_{\mathcal{N}(v_i)}) - f( \mathcal{G}_{\mathcal{N} (v_i) \setminus \mathcal{N}^*_{v_i}} ) \triangleq \mathrm{Fidelity}^+(v_i, \mathcal{N}^*_{v_i})$ represents the $\mathrm{Fidelity}^+$ of the explanatory nodes $\mathcal{N}^*_{v_i}$ for a single node $v_i$; and $N$ refers to the number of nodes in the graph.
Higher $\mathrm{Fidelity}^+$ indicates a better explanation, because the discriminative neighbors of each node are recognized and removed, resulting in huge prediction changes.  


On the other hand, $\mathrm{Fidelity}^-$ evaluates the GNN explanation via \textit{removing the trivial neighbors} of each node. Thus, $\mathrm{Fidelity}^-$ is formalized as follows:
\begin{align}
 \downarrow  \mathrm{Fidelity}^- = \frac{1}{N}\sum_{i=1}^N f(\mathcal{G}_{\mathcal{N}(v_i)}) - f(\mathcal{G}_{\mathcal{N}^*_{v_i}}),
 \label{eq:fidelity-}
\end{align}
where $f(\mathcal{G}_{\mathcal{N}(v_i)}) - f(\mathcal{G}_{\mathcal{N}^*_{v_i}}) \triangleq \mathrm{Fidelity}^-(v_i, \mathcal{N}^*_{v_i})$ refers to the $\mathrm{Fidelity}^-$ for a single node $v_i$.
Lower $\mathrm{Fidelity}^-$ implies a better explanation, since the contributive neighbors of each node have been preserved to keep the prediction similar to the original one. 



\paragraph{Throughput.} Following existing work~\cite{chuang2022cortx,wang2022accelerating, liu2022exact}, we utilize algorithmic throughput to evaluate the efficiency of explanation methods. 
Specifically, the throughput is defined as
$N_{\text{test}}/{t_{\text{test}}}$, where $N_{\text{test}}$ is the number of instances being explained, and $t_{\text{test}}$ represents the total time cost of the explanation process. 
Thus, the throughput indicates the number of instances being explained per second, and a higher throughput implies a more efficient explanation method. 


%% file: text/3-removal-based-explanation.tex
\section{Removal-based Attribution}
\label{sec:removal-based-attribution}

\begin{figure}[b]
     \centering
     \begin{subfigure}[b]{0.32\textwidth}
         \centering
         \includegraphics[width=\textwidth]{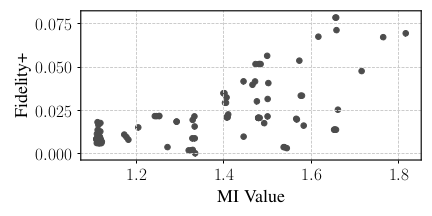}
         \caption{}
         \label{fig:teaser-a}
     \end{subfigure}
     \hfill
     \begin{subfigure}[b]{0.32\textwidth}
         \centering
         \includegraphics[width=\textwidth]{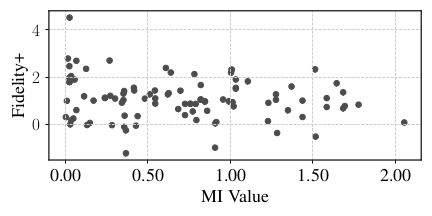}
         \caption{}
        \label{fig:teaser-b}
     \end{subfigure}
     \hfill
     \begin{subfigure}[b]{0.32\textwidth}
         \centering
         \includegraphics[width=\textwidth]{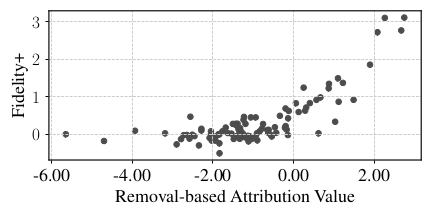}
         \caption{}
         \label{fig:teaser-c}
     \end{subfigure}
        \caption{\footnotesize (a) Correlation between fidelity and mutual information value on the BA-Community and (b) ogbn-arxiv datasets; (c) Correlation between fidelity and removal-based attribution on the ogbn-arixv dataset.}
        \label{fig:teaser}
\end{figure}

In this section, we first demonstrate the limitation of using mutual information (MI) as guidance for faithful GNN explanation generation in existing methods.
Then, we define removal-based attribution and theoretically demonstrate its consistency with the expected interpretation fidelity.
Finally, we experimentally show the correlation between fidelity and removal-based attribution.

Many existing work trains an explainer by a loss adopted from MI (e.g., MI with regularization~\cite{luo2020parameterized} or the causal counterpart of MI~\cite{lin2022orphicx}), so that the trained explainer provides explanatory features (e.g., edge importance)  based on the MI value in the objective function.
However, we found the MI value becomes less consistent with the interpretation fidelity as the number of neighbor nodes increases. 
To illustrate this issue, we conduct a preliminary experiment on the dataset BA-Community~\cite{ying2019gnnexplainer} and ogbn-arxiv~\cite{hu2020open}, and visualize the correlation of the interpretation fidelity with the MI value for the target nodes with different numbers of neighbors in \Cref{fig:teaser}.
The details about the experiment are given in \Cref{supp:removal-based attribution experiment}.
It is observed in \Cref{fig:teaser}~(a) that the MI value is approximately correlated with fidelity as the number of neighbors is small. However, as the number of neighbors grows on the dataset obgn-arxiv, this correlation becomes less evident in \Cref{fig:teaser}~(b), indicating the unfaithfulness of using MI value as explanations on larger graphs.

To address the limitation of MI value, we propose removal-based attribution that consistently reflects fidelity.
Specifically, we combine two fidelity metrics $\mathrm{Fidelity}^+$ and $\mathrm{Fidelity}^-$ given by Equations~(\ref{eq:fidelity+}) and~(\ref{eq:fidelity-}) into a hybrid formulation of fidelity as follows: 
\begin{align}
\label{eq:delta_fidelity}
    \Delta \mathrm{Fidelity}(v_i, \mathcal{N}^*_{v_i}) &= \mathrm{Fidelity}^+(v_i, \mathcal{N}^*_{v_i}) - \mathrm{Fidelity}^-(v_i, \mathcal{N}^*_{v_i}).
\end{align}

Note that the estimation in \Cref{eq:delta_fidelity} is considered for a neighbor node set $\mathcal{N}^*_{v_i}$.
We extend \Cref{eq:delta_fidelity} to the case of each single neighbor node $v_j \!\in\! \mathcal{N}(v_i)$ by considering the expected fidelity $\mathbb{E}_{\mathcal{S}_j \subseteq \mathcal{N}(v_i)}[\Delta \mathrm{Fidelity}(v_i, \mathcal{S}_j)]$, where the enumeration of $\mathcal{S}_j \!\subseteq\! \mathcal{N} (v_i)$ satisfies $\mathcal{S}_j \!\ni\! v_j$.
Without loss of generality, we give Theorem~\ref{theorem:removal-based-attribution} to demonstrate the consistency of expected fidelity with the expected GNN prediction difference. The proof of Theorem~\ref{theorem:removal-based-attribution} is given in \Cref{supp:theorem proof}.
\begin{theorem} 
\label{theorem:removal-based-attribution}
For a target node $v_i$ and one of its neighbors $v_j$, let $\phi_{j \to i} =  \mathbb{E}_{\mathcal{S}_j \subseteq \mathcal{N}(v_i)} \big[ f(\mathcal{G}_{\mathcal{S}_j}) - f(\mathcal{G}_{\mathcal{N}({v_i)} \setminus \mathcal{S}_j }) \big]$. Generally, $\forall \mathcal{J} \!\subseteq\! \mathcal{N}(v_i)$, if $\exists \mathcal{J}^* \!\subseteq\! \mathcal{N}(v_i)$ satisfying $\mathbb{E}_{j \sim \mathcal{J}} \phi_{j \to i} \leq \mathbb{E}_{j \sim \mathcal{J}^*} \phi_{j \to i}$, then
    \begin{align}
        \mathbb{E}_{j \sim \mathcal{J}} \mathbb{E}_{\mathcal{S}_j \subseteq \mathcal{N}(v_i)}[\Delta \mathrm{Fidelity}(v_i, \mathcal{S}_j)] \leq \mathbb{E}_{j \sim \mathcal{J}^*} \mathbb{E}_{\mathcal{S}_j \subseteq \mathcal{N}(v_i)}[\Delta \mathrm{Fidelity}(v_i, \mathcal{S}_j)].
    \end{align}
\end{theorem}
According to Theorem~\ref{theorem:removal-based-attribution}, if the expected prediction difference of a node subset $\mathcal{J}^*$ is larger than that of any node subset $\mathcal{J}$, then the expected fidelity of the node subset $\mathcal{J}^*$ is also larger than that of any node subset.
As we aim at a node subset that maximizes the fidelity, the objective can be altered to find such a node subset $\mathcal{J}^*$ for the target node $v_i$ through $\mathbb{E}_{j \sim \mathcal{J}} \phi_{j \to i}$.

Formally, we define $\phi_{j \to i}$ as the \textit{removal-based attribution} for each neighbor $v_j$ of the target node $v_i$ to indicate its importance to the GNN prediction $f(\mathcal{G}_{\mathcal{N}(v_i)})$ in Definition~\ref{define:removal-based-attribution}.
\begin{definition}[Removal-based Attribution] 
\label{define:removal-based-attribution}
For a target node $v_i$ and neighbor node $v_j$, the removal-based attribution $\phi_{j \to i}$ is defined as
\begin{align}
\label{eq:removal-based-attribution}
\phi_{j \to i} =  \mathbb{E}_{\mathcal{S}_j \subseteq \mathcal{N}(v_i)} \big[ f(\mathcal{G}_{\mathcal{S}_j}) - f(\mathcal{G}_{\mathcal{N}({v_i)} \setminus \mathcal{S}_j }) \big] = \frac{1}{2^{|\mathcal{N}(v_i)|-1}} \!\!\!\!\!\! \sum_{\mathcal{S}_j \subseteq \mathcal{N}(v_i)} f(\mathcal{G}_{\mathcal{S}_j}) - f(\mathcal{G}_{\mathcal{N}({v_i)} \setminus \mathcal{S}_j }).
\end{align}
\end{definition}

According to the numerical result in Figure~\ref{fig:teaser}~(c).
A stronger correlation can be observed between the interpretation fidelity with our proposed removal-based attribution.
This indicates the advantage of utilizing removal-based attribution as node attribution in GNN explanations.

%% file: text/4-methodology.tex
\section{\Algtitle{}~(\Algnameabbr{})}
\label{sec:method}

\begin{figure*}[t]
\centering
\includegraphics[width=.99\linewidth]{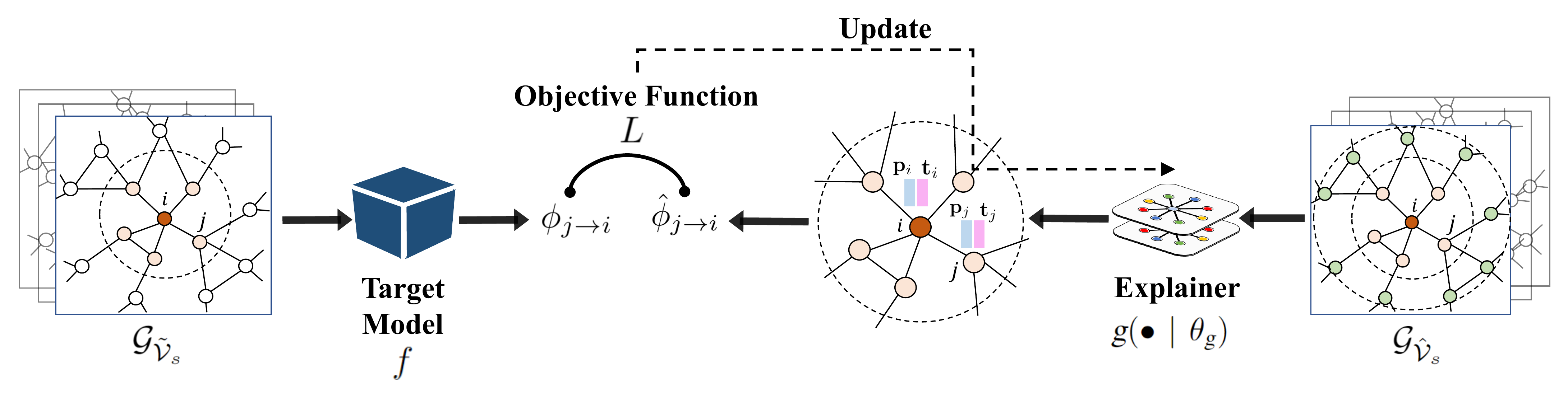}
\vspace{-5pt}
\caption{The framework of \Algnameabbr{}. \textbf{Left}: Subgraphs are sampled to estimate the removal-based attribution $\phi_{j\to i}$. \textbf{Right}: $\textsl{g}(\bullet ~|~ \theta_{\textsl{g}})$ samples subgraphs as input and generates the node attribution $\hat{\phi}_{j\to i}$ based on the source embedding $\mathbf{p}_j$ and target embedding $\mathbf{t}_i$. \textbf{Middle}: The parameters of the amortized explainer are updated via optimizing the training objective given by Equation~(\ref{eq:loss_func}).}
\label{fig:method}
\end{figure*}

We introduce details of the proposed \Algnameabbr{} framework in this section.
As illustrated in Figure~\ref{fig:method}, \Algnameabbr{} learns a GNN-based explainer on a training dataset, where the details of explainer training and training sample collection are given in \Cref{sec:explaienr}, and \Cref{sec:subsample}, respectively.


\subsection{Explainer Training}
\label{sec:explaienr}
We introduce how the amortized explainer in \Algnameabbr{} generates the removal-based attribution in this section.
Different from existing work of amortized explainers on the regular data~\cite{jethani2021fastshap, covert2021explaining, chuang2022cortx}, graphs are irregular and non-Euclidean data, where the number of nodes and edges vary in different graph regions, and each node may have a different number of neighbors.
\Algnameabbr{} adopts a GNN-based amortized explainer $\textsl{g}(\bullet ~|~ \theta_{\textsl{g}})$ to to solve this problem.
Specifically, for a node $v_i \!\in\! \mathcal{G}$, $\textsl{g}(\bullet ~|~ \theta_{\textsl{g}})$ generates explanation-oriented embeddings $[\mathbf{p}_i, \mathbf{t}_i] = \textsl{g}((\mathcal{N}(v_i), \mathcal{E}_{\mathcal{N}(v_i)})  ~|~ \theta_{\textsl{g}})$, where $\mathbf{p}_i \in \mathbb{R}^n$ and $\mathbf{t}_i \in \mathbb{R}^n$ denote the \emph{source} and \emph{target} embedding of node $v_i$, respectively, and the embedding dimension $n$ is a hyper-parameter.
The source embedding $\mathbf{p}_i$ indicates the attribution of $v_i$~(as a source node) to other nodes, and target embedding is used to compute the attribution of other nodes to this node $v_i$~(as a target node). 
In this way, the attribution of a source node $v_j$ to a target node can be estimated by an inner product of the source embedding $\mathbf{p}_j$ and target embedding $\mathbf{t}_i$ as follows:
\begin{equation}
\label{eq:phi_hat}
    \hat{\phi}_{j \to i} = \langle \mathbf{p}_j, \mathbf{t}_i \rangle,
\end{equation}
where $\hat{\phi}_{j \to i}$ is the estimated attribution of source node $v_j$ to target node $v_i$.
Notably, the removal-based attribution is asymmetric according to \Cref{eq:removal-based-attribution}, i.e. $\phi_{j \to i} \neq \phi_{i \to j}$, as a result of asymmetrical information flows within the graph structure. Correspondingly, our framework generates different attributions when a node is a source or a target, i.e. $\langle \mathbf{p}_j, \mathbf{t}_i \rangle \neq \langle \mathbf{p}_i, \mathbf{t}_j \rangle$ for $i \neq j$.
Likewise, the attribution score of a source node $v_j$ to a target graph/subgraph $\mathcal{G}$ ($v_j \in \mathcal{G}$) can be defined as follows: 
\begin{equation}
\label{eq:phi_hat_graph}
    \hat{\phi}_{j \to \mathcal{G}} = \langle \mathbf{p}_j, \mathbf{t}_{\mathcal{G}} \rangle,
\end{equation}
where $\hat{\phi}_{j \to \mathcal{G}}$ denotes the attribution of node $v_j$ to a target graph $\mathcal{G}$; and $\mathbf{t}_{\mathcal{G}} \in \mathbb{R}^n$ is the target embedding of $\mathcal{G}$, which takes the pooling of target embedding $\mathbf{t}_i$ for nodes $v_i \!\in\! \mathcal{G}$.

The parameters of explainer $\textsl{g}(\bullet ~|~ \theta_{\textsl{g}})$ are trained to learn the removal-based attribution $\phi_{j \to i}$ introduced in \Cref{eq:removal-based-attribution} by minimizing the following loss:
\begin{equation}
\label{eq:loss_func}
L = \mathbb{E}_{v_i \sim \mathcal{V}, v_j \sim \mathcal{N} (v_i)} \Big[ ( \hat{\phi}_{j \to i} - \phi_{j \to i} )^2\Big].
\end{equation}
Note that the complexity of \Cref{eq:removal-based-attribution} is exceedingly high.
We incorporate the Monte Carlo estimation of $\phi_{j \to i}$ as an approximation into Equation~(\ref{eq:loss_func}), which is given as follows:
\begin{equation}
    \label{eq:tilde_phi_sample}
    \phi_{j \to i} \approx \mathbb{E}_{\mathcal{S}_j \sim \mathcal{N}(v_i)} \big[ f(\mathcal{G}_{\mathcal{S}_j}) \!-\! f(\mathcal{G}_{\mathcal{N}(v_i) \setminus \mathcal{S}_j}) \big].
\end{equation}
Moreover, Equation~(\ref{eq:tilde_phi_sample}) can be re-formulated into the recursive process as follows,
\begin{gather}
\label{eq:tilde_phi_iteration}
    \tilde\phi_{j \to i}^{(t)} \!=\! (1 \!-\! t^{-1}) \tilde{\phi}_{j \to i}^{(t-1)} \!+\! t^{-1} \! \big[ f(\mathcal{G}_{\mathcal{S}_j}) \!-\! f(\mathcal{G}_{\mathcal{N}(v_i) \setminus \mathcal{S}_j}) \big] \!, \!
\end{gather}
where $\mathcal{S}_j \!\sim\! \mathcal{N}(v_i), \mathcal{S}_j \ni v_j$ denotes a subset of nodes uniformly sampled from the neighbors $\mathcal{N}(v_i)$;
$\tilde{\phi}_{j \to i}^{(t)}$ denotes the estimation of $\phi_{j \to i}$ in epoch $t$; and its initial value in the first epoch takes $\tilde{\phi}_{j \to i}^{(0)} \!=\! 0$. 
To this end, the estimation of ${\phi}_{j \to i}$ can be integrated with the update of the explainer parameter $\theta_{\textsl{g}}$ into an iterative process, where a hard threshold for subgraph sampling is no longer required.

After the learning of explainer $\textsl{g}(\bullet ~|~ \theta_{\textsl{g}})$, for each target node $v_i$ or graph $\mathcal{G}$, the importance of each source node can be generated following the inference process given by Equations~(\ref{eq:phi_hat}) or (\ref{eq:phi_hat_graph}), respectively. As our explainer is GNN-based, it can be applied to complex graph structures and generate the attribution for each node in an inductive way, once it is trained.

\subsection{Subgraph Sampling}
\label{sec:subsample}
Note that the computational unit~(GPU) has limited memory capacity to store the graph during the training.
We design a subgraph sampling strategy guided by the training process of the amortized explainer for handling large graph datasets.
Specifically,  in each iteration, \Algnameabbr{} samples a mini-batch of target nodes $\mathcal{V}_t \subseteq \mathcal{V}$.
According to the message passing in GNNs, the reception field of a node depends on the number of aggregation layers, which is defined as $K$. 
Thus, \Algnameabbr{} samples the $K$-hop neighbors of the target nodes to construct the subgraph $\mathcal{G}_{\tilde{\mathcal{V}}_s}$ to compute the attribution $\tilde{\phi}_{j \to i}^{(t)}$.
Formally, the neighbor samples are given by
\begin{equation}
\label{eq:subgraph1}
\tilde{\mathcal{V}}_s = \{ v_1, v_2 ~|~ d(v_1, v_2) \leq K, v_1 \in \mathcal{V}_t, v_2 \in \mathcal{V} \}.
\end{equation}
After that, \Algnameabbr{} samples another subgraph for updating the parameters of the explainer.
According to Equation~(\ref{eq:phi_hat}), the deduction of explanation-oriented embedding $\mathbf{t}_i$ for $v_i \in \mathcal{V}_s$ requires the $K$-hop neighbors of the target nodes, so $\mathbf{p}_j$ for $v_j \in \mathcal{N}(v_i)$ requires $2K$-hop neighbors in total.
Therefore, \Algnameabbr{} takes a subgraph $\mathcal{G}_{\hat{\mathcal{V}}_s}$ that consists of $2K$-hop neighbors of the target nodes and the related edges.
In this way, the sampled nodes $\hat{\mathcal{V}}_s$ are given by
\begin{equation}
\label{eq:subgraph2}
 \hat{\mathcal{V}}_s = \{ v_1, v_2 ~|~ d(v_1, v_2) \leq 2K, v_1 \in \mathcal{V}_t, v_2 \in \mathcal{V}\}.
\end{equation}
The subgraph $\mathcal{G}_{\hat{\mathcal{V}}_s}$ is sampled from the whole graph $\mathcal{G}$ for updating the parameters of the explainer by minimizing the loss in \Cref{eq:loss_func}.

\begin{wrapfigure}{R}{0.5\textwidth}
  \scalebox{0.95}{
    \begin{minipage}{\linewidth}
        \vspace{-23pt}
        \begin{algorithm}[H]
        \footnotesize
           \caption{Training of Explainer $\textsl{g}(\bullet ~|~ \theta_{\textsl{g}}^*)$.}
           \label{alg:regex}
        \begin{algorithmic}[1]
           \STATE {\bfseries Input:} Graph $\mathcal{G}(\mathcal{V}, \mathcal{E})$, target GNN model $f(\cdot)$.
           \STATE {\bfseries Output:} Explainer $\textsl{g}(\bullet ~|~ \theta_{\textsl{g}}^*)$.
           \STATE Initialize $\tilde{\phi}_{j \to i}^{(0)} \!=\! 0$ for $v_i, v_j \!\in\! \mathcal{G}$. Set $t = 1$
           \WHILE{\emph{not converge}}
           \STATE {\color{lightgray} \# Estimate the removal-based attribution.}
           \STATE Sample a mini-batch of target nodes $\mathcal{V}_t \subseteq \mathcal{V}$.
           \STATE Sample a subgraph following Equation~(\ref{eq:subgraph1}).
           \STATE Update $\tilde{\phi}_{j \to i}^{(t)}$ following Equation~(\ref{eq:tilde_phi_iteration}).
           \STATE {\color{lightgray} \# Update explainer parameters.}
           \STATE Sample a subgraph following Equation~(\ref{eq:subgraph2}).
           \STATE Update the parameters of $\textsl{g}(\bullet ~|~ \theta_{\textsl{g}})$ by
           \begin{equation}
           \vspace{-2mm}
               \theta_{\textsl{g}}^* \!=\! \arg \min_{\theta} \mathbb{E}_{v_i \sim \mathcal{V}_t, v_j \sim \mathcal{N} (v_i)} \! \Big[ \! ( \hat{\phi}_{j \to i} - \tilde{\phi}_{j \to i}^{(t)})^2 \! \Big]
               \nonumber
           \vspace{-2mm}
           \end{equation}
           \STATE $t = t + 1$
           \ENDWHILE
        \end{algorithmic}
        \end{algorithm}
    \end{minipage}
  }
  \vspace{-17pt}
\end{wrapfigure}
  
The training of amortized explainer with subgraph sampling is summarized in Algorithm~\ref{alg:regex}.
To be concrete, \Algnameabbr{} first samples a subgraph~(lines 6-7); then update the estimation of removal-based attribution for the source nodes~(line 8); finally update the parameters of the explainer ~(line 11).
The iteration ends with the convergence of the amortized explainer.

\paragraph{Estimation of the Max Hop.}
\label{sec:max hop}
The value of $K$ in sampling is directly related to the maximum hop in the message passing. However, in black-box scenarios, we do not know the number of aggregation layers in the target GNN model. 
A straightforward solution to this problem is to search for the maximum hop neighbors that can disturb the output of the target GNN model.
Concretely, let $\mathcal{G}_{\mathcal{N}_k(v)}$ denote the nodes within the $k$-hop neighbors of a node $v$.
The maximum hop of the target GNN model is the minimum $k$ that satisfies $f(\mathcal{G}_{\mathcal{N}_{k+1}(v)}) = f(\mathcal{G}_{\mathcal{N}_k(v)})$ as follows, 
\begin{equation}
\label{eq:max_hop}
    K = \min \{ k~|~ f(\mathcal{G}_{\mathcal{N}_{k+1}(v)}) = f(\mathcal{G}_{\mathcal{N}_k(v)}) \},
\end{equation}
where node $v$ can be randomly sampled from the graph. 
Since a practically deployed GNN model usually cannot have too many aggregation layers to avoid the problem of over-smoothing, the solution of the above problem can be achieved via a grid search over $k = 1, 2, \cdots$.

%% file: text/5-experiments.tex
\section{Experiment}

In this section, we conduct experiments to demonstrate the effectiveness of \Algnameabbr{} in explaining node classification and graph classification results. 
We quantitatively evaluate \Algnameabbr{} with respect to interpretation fidelity and throughput in \Cref{sec:quanti res}, followed by ablation study and hyper-parameter analysis for \Algnameabbr{}. \Cref{sec:quali res} demonstrates several case studies on node and graph classification.

\begin{figure*}[t]
     \centering
     \begin{subfigure}[b]{0.32\textwidth}
         \centering
         \includegraphics[width=\textwidth]{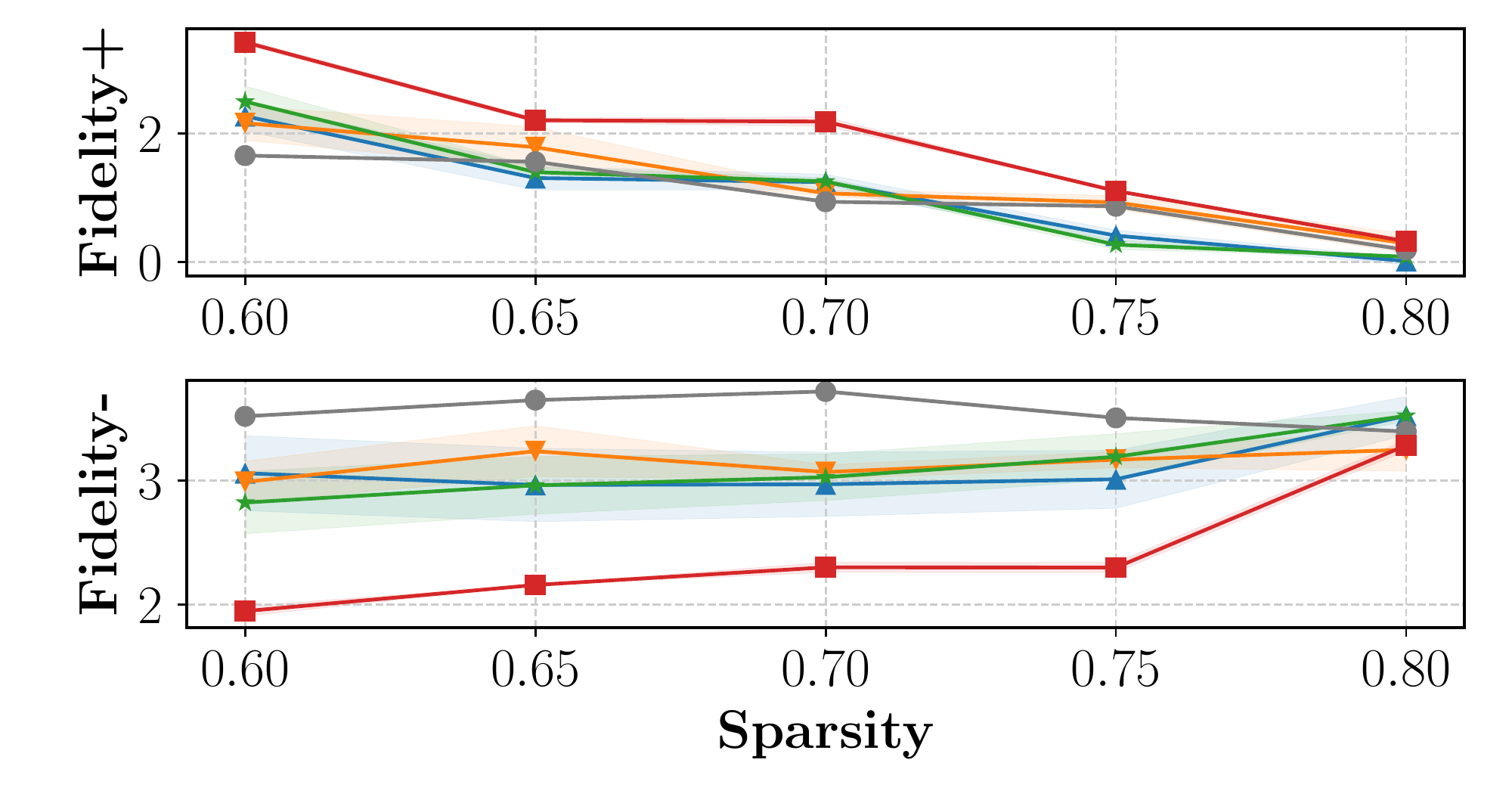}
         \caption{BA-Shapes}
         \label{subfig:f-bashapes}
     \end{subfigure}
     \hfill
     \begin{subfigure}[b]{0.32\textwidth}
         \centering
         \includegraphics[width=\textwidth]{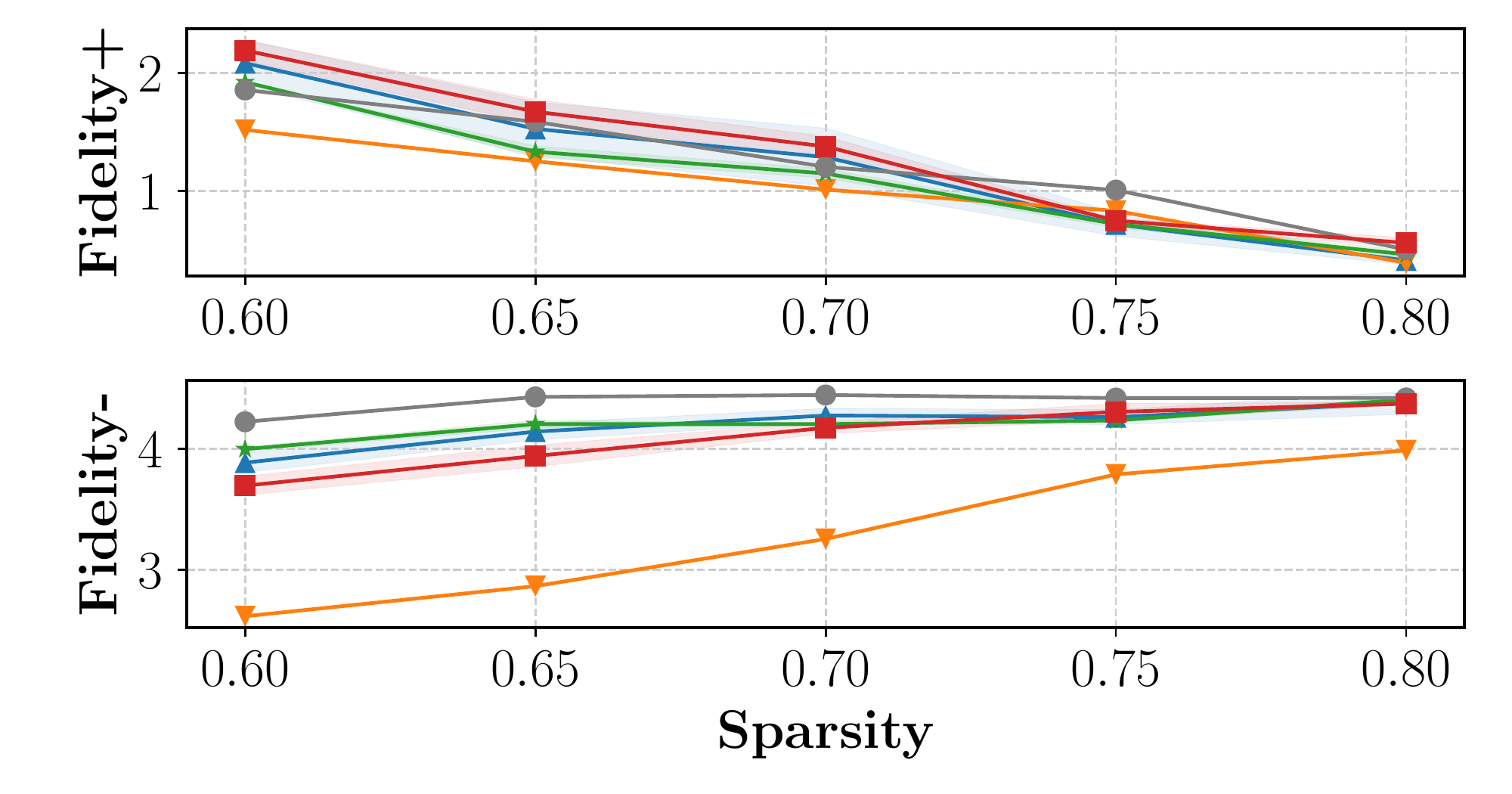}
         \caption{BA-Community}
        \label{subfig:f-bacomm}
     \end{subfigure}
     \hfill
     \begin{subfigure}[b]{0.32\textwidth}
         \centering
         \includegraphics[width=\textwidth]{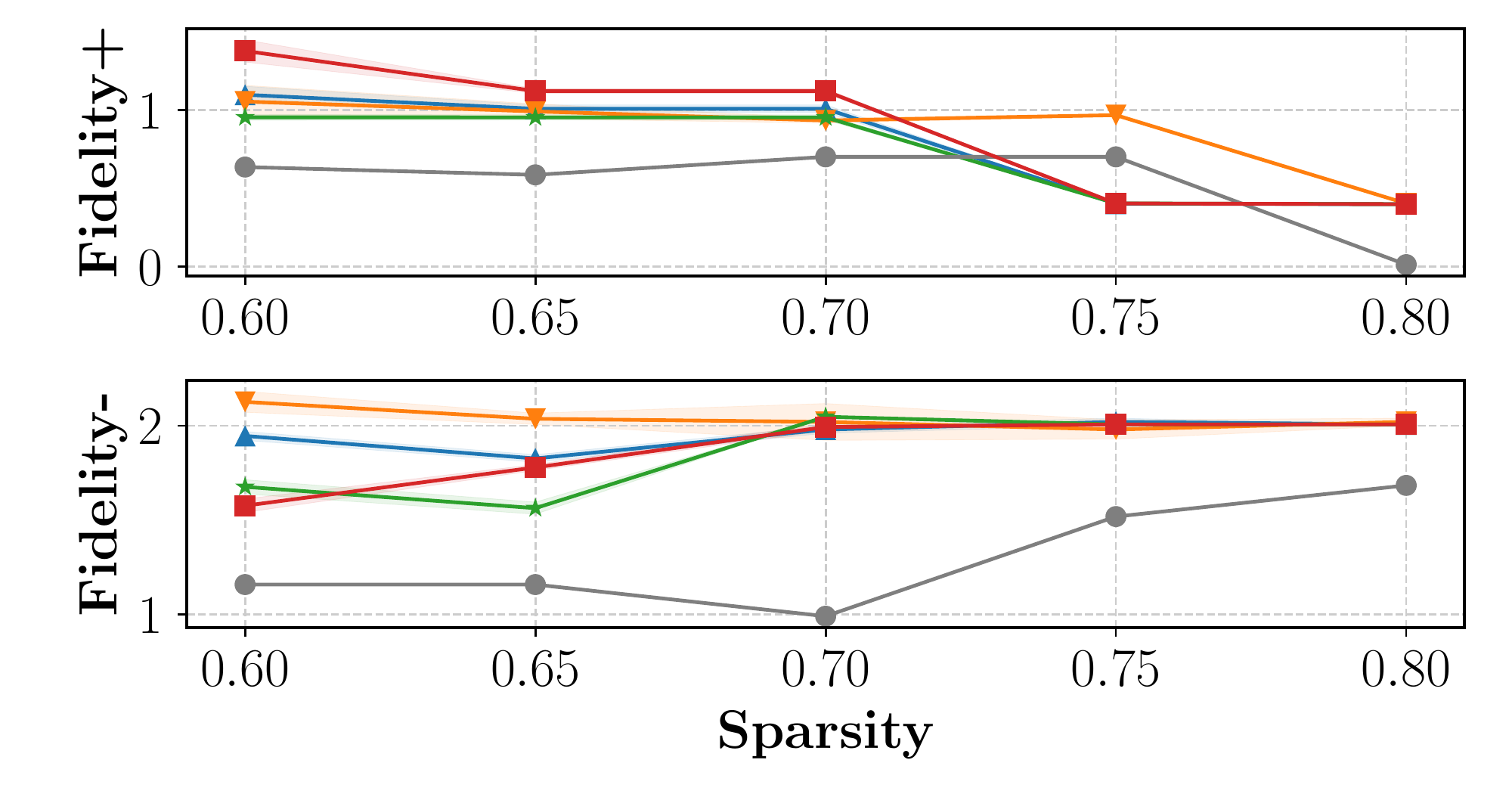}
         \caption{Tree-Cycles}
         \label{subfig:f-tree}
     \end{subfigure}
     \\
     \begin{subfigure}[b]{0.32\textwidth}
         \centering
         \includegraphics[width=\textwidth]{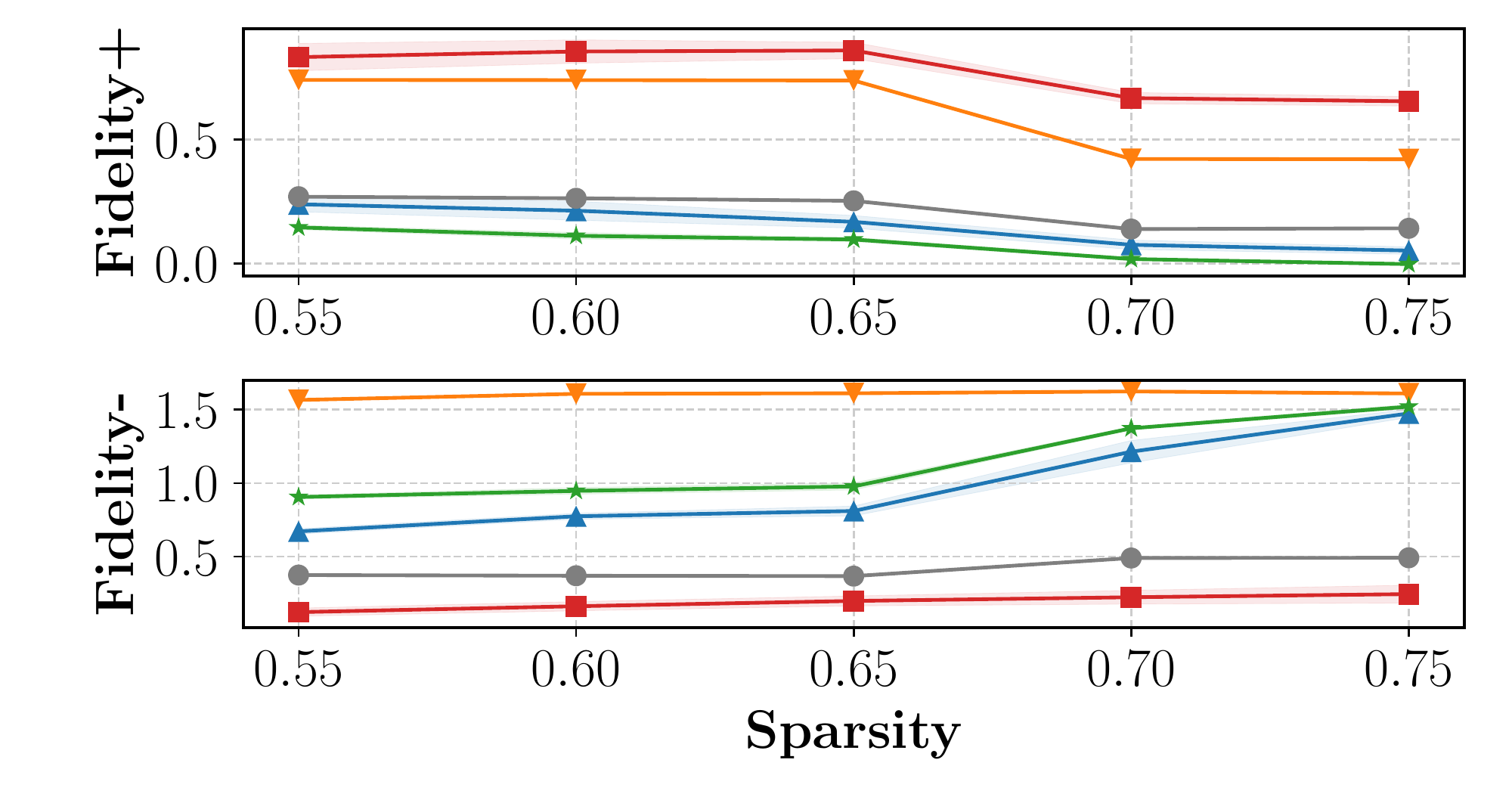}
         \caption{ogbn-arxiv}
         \label{subfig:f-arxiv}
     \end{subfigure}
     \hfill
    \begin{subfigure}[b]{0.32\textwidth}
         \centering
         \includegraphics[width=\textwidth]{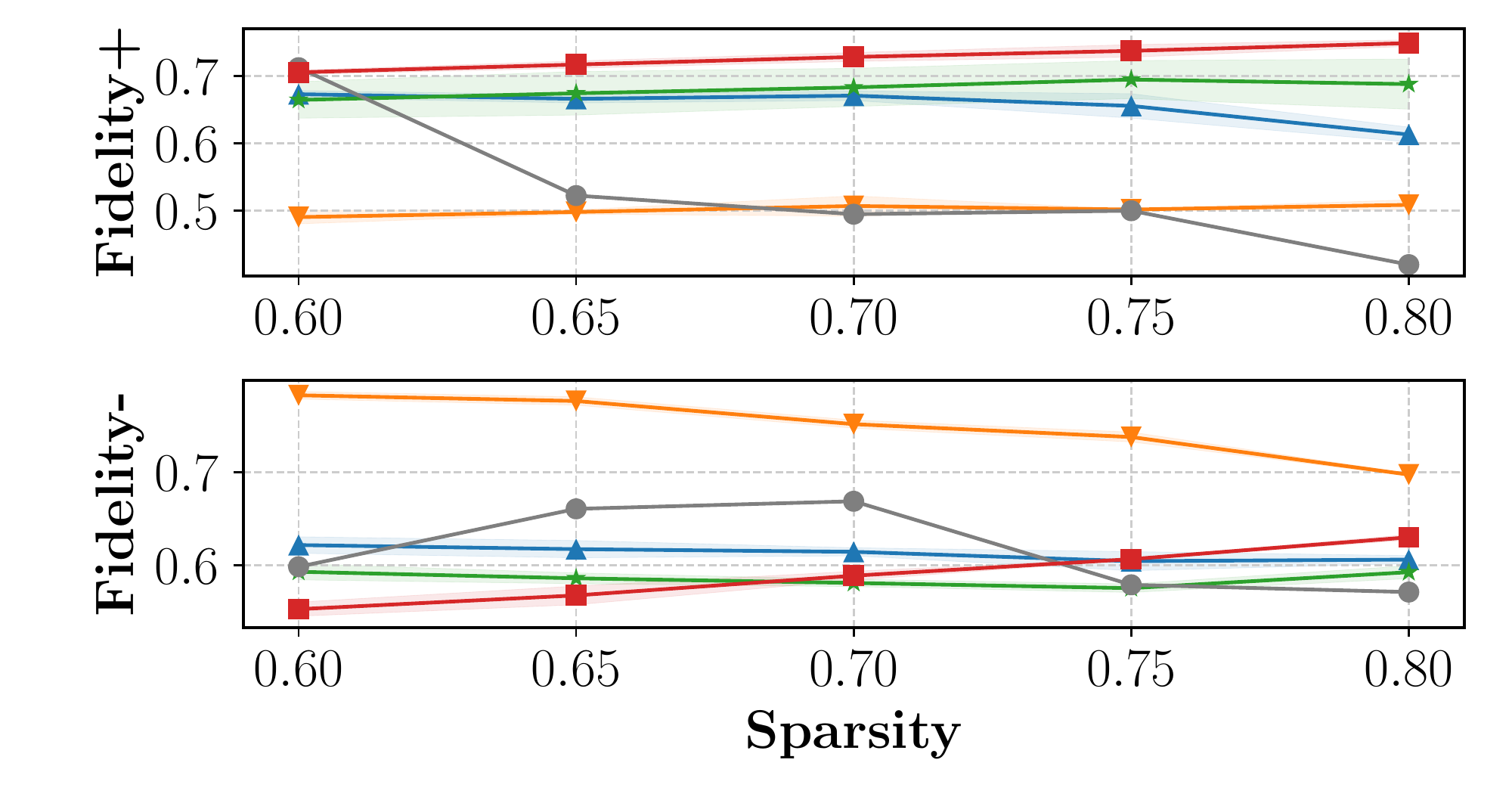}
         \caption{BA2Motifs}
         \label{subfig:f-ba2}
     \end{subfigure}
     \hfill
     \begin{subfigure}[b]{0.32\textwidth}
         \centering
         \includegraphics[width=.9\textwidth]{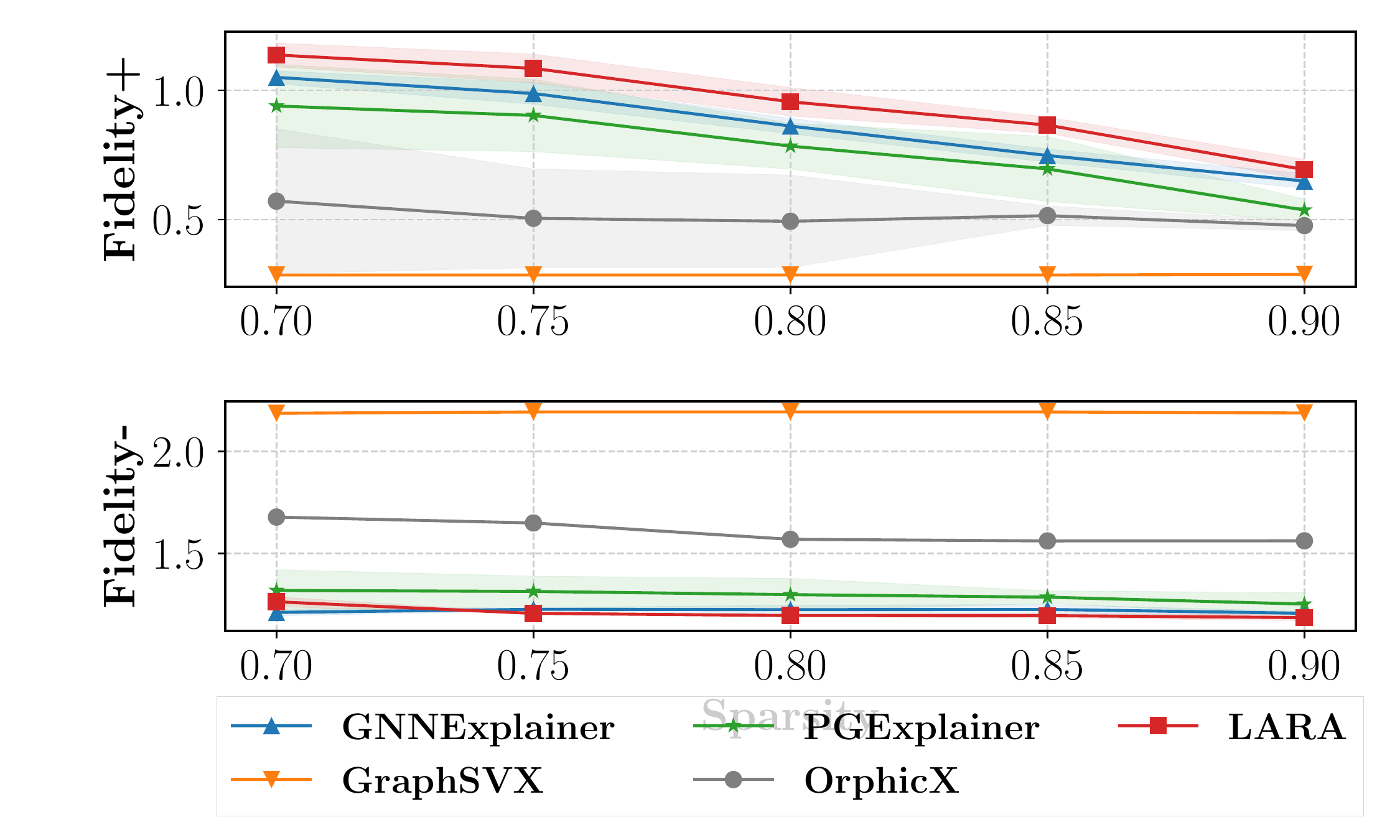}
         \caption{MUTAG}
         \label{subfig:f-mutag}
     \end{subfigure}
    \caption{\footnotesize Interpretation fidelity versus sparsity of neighbors on (a) BA-Shape, (b) BA-Community, (c) Tree-Cycles, (d) ogbn-arxiv, (e) BA2Motifs, and (f) MUTAG datasets. In sub-figures of $\mathrm{Fidelity^+}$, a line placed above indicates better performance. In sub-figures of $\mathrm{Fidelity^-}$, a line placed below indicates better performance.}
    \label{fig:fidelity}
\end{figure*}

\subsection{Quantitative Evaluation}
\label{sec:quanti res}
We evaluate the faithfulness and efficiency of \Algnameabbr{} compared to state-of-the-art methods:
GNNExplainer~\cite{ying2019gnnexplainer}, PGExplainer~\cite{luo2020parameterized}, GraphSVX~\cite{duval2021graphsvx} and OrphicX~\cite{lin2022orphicx}. The experiments are conducted on six benchmark datasets: BA-Shapes, BA-Community, Tree-Cycles~\cite{ying2019gnnexplainer} and ogbn-arxiv~\cite{hu2020open} for node classification, and BA-2Motifs~\cite{luo2020parameterized} and MUTAG~\cite{debnath1991structure} for graph classification. More details about the implementation, datasets, baseline methods, and hyper-parameter settings are given in~\Cref{supp:experiments}.

\begin{table*}[t]
\caption{\label{tab:throughput} Throughput of \Algnameabbr{} compared with state-of-the-art methods. Results are shown in (mean $\pm$ std) over five runs and the best results are marked in bold.}
\centering
\resizebox{\textwidth}{!}{
\begin{tabular}{l|c|c|c|c|c|c|c}
\toprule[1pt]
& Method & BA-Shapes & BA-Community & Tree-Cycles & BA-2Motifs & MUTAG & ogbn-arxiv \\
\hline
\multirow{5}{*}{Throughput $\uparrow$} & GNNExplainer & 3.15$\pm$\scriptsize{0.11} & 2.19$\pm$\scriptsize{0.09} & 4.18$\pm$\scriptsize{0.03} & 6.05$\pm$\scriptsize{0.02} & 15.38$\pm$\scriptsize{0.63} & 9.35$\pm$\scriptsize{0.06} \\
& GraphSVX & 9.51$\pm$\scriptsize{0.04} & 10.58$\pm$\scriptsize{0.67} & 10.31$\pm$\scriptsize{0.19} & 4.90$\pm$\scriptsize{0.18} & 2.09$\pm$\scriptsize{0.21} & 0.95$\pm$\scriptsize{0.06} \\
& PGExplainer & 22.95$\pm$\scriptsize{0.14} & 13.67$\pm$\scriptsize{0.18} & 423.78$\pm$\scriptsize{4.51} & 490.55$\pm$\scriptsize{9.94} & 80.39$\pm$\scriptsize{2.53} & 28.38$\pm$\scriptsize{0.07} \\
& OrphicX & 262.49$\pm$\scriptsize{7.95} & 196.23$\pm$\scriptsize{3.39} & 447.39$\pm$\scriptsize{5.10} & 373.21$\pm$\scriptsize{5.12} & 286.07$\pm$\scriptsize{3.07} & 77.85$\pm$\scriptsize{0.07} \\
& \Algnameabbr{} & \textbf{670.54}$\pm$\scriptsize{1.94} & \textbf{644.63}$\pm$\scriptsize{1.03} & \textbf{674.12}$\pm$\scriptsize{4.06} & \textbf{635.10}$\pm$\scriptsize{7.24} & \textbf{334.44}$\pm$\scriptsize{2.18} & \textbf{237.97}$\pm$\scriptsize{3.87} \\
\bottomrule[1pt]
\end{tabular}
}
\vspace{-10pt}
\end{table*}

\vspace{-5pt}
\paragraph{Fidelity.}
\Cref{fig:fidelity} shows the $\mathrm{Fidelity}^+$ and $\mathrm{Fidelity}^-$ versus the sparsity of neighbors\footnote{\scriptsize In the experiment of $\mathrm{Fidelity}^+$, a sparsity of $p\%$ indicates top~$(1-p\%)$ important neighbors are removed; in the experiment of $\mathrm{Fidelity}^-$, a sparsity of $p\%$ indicates the least important~$p\%$ neighbors are removed, following the definition in~\cite{yuan2022explainability}.}, where \Algnameabbr{} is compared with state-of-the-art baseline methods on the six benchmark datasets. 
Overall, \Algnameabbr{} (in red) outperforms baseline methods on most of the datasets in terms of higher $\mathrm{Fidelity}^+$ and lower $\mathrm{Fidelity}^-$, particularly on the large graph dataset ogbn-arxiv, as shown in \Cref{fig:fidelity}~(d). This highlights its potential in faithful explanations for real-world use cases. Among the baseline methods, GNNExplainer, PGExplainer and OrphicX adopt the MI value for node attribution, which is less effective than \Algnameabbr{}, especially on ogbn-arxiv. This is consistent with our observations in Section~\ref{sec:removal-based-attribution}, demonstrating the advantage of \Algnameabbr{} towards maximizing the interpretation fidelity by learning our proposed removal-based attribution.


\vspace{-7pt}
\paragraph{Efficiency.}
According to \Cref{tab:throughput}, \Algnameabbr{} shows a significant improvement in throughput on all datasets compared with baseline methods.
As a non-amortized method, GNNExplainer shows a considerably lower throughput of interpretation than other methods, because it has to learn an explainer for each instance. GraphSVX relies on solving a matrix equation to estimate the importance of each neighbor. This is separately conducted for each target node, resulting in considerably low efficiency.
Although PGExplainer and OrphicX can simultaneously generate a batch of explanations, their time complexity increases with the edge number contained in the relevant subgraph of the target node.
As an amortized method, \Algnameabbr{} has a constant time complexity of the explanation based on a feed-forward process of the explainer, and can efficiently explain multiple nodes within a batch.

\vspace{-7pt}
\paragraph{Scalability.}
We study the scalability of \Algnameabbr{} on a large graph ogbn-arxiv. The time latency of explaining different numbers of nodes from the ogbn-arxiv are listed in \Cref{tab:time_latency_arxiv}.
In particular, as the node number grows from $10^2$ to $10^5$, GNNexplainer and PGExplainer show approximate $12000$s and $3500$s growth of time latency~($\Delta$~time~latency), respectively.
In contrast, \Algnameabbr{} has only $400$s growing, which shows stronger scalability than baseline methods. Moreover, \Algnameabbr{} exhibits about $90\%$ less time latency compared to the competitive baseline PGExplainer in general.
The result demonstrates the potential of \Algnameabbr{} in real-world applications such as in a social network with millions of nodes and edges.

\vspace{-5pt}
\subsection{Ablation Study and Hyper-parameter Analysis}
In the ablation study, we validate the effectiveness of key components in the \Algnameabbr{} framework: 1) learning the removal-based attribution; 2) bidirectional embedding~(source and target embeddings are learned separately).
After this, we study the interpretation fidelity affected by the key hyper-parameter of \Algnameabbr{}, i.e., the number of aggregation layers in the amortized explainer.

\vspace{-6pt}
\paragraph{Learning Removal- v.s. MI-based Attribution.}
We demonstrate the advantage of learning removal-based attribution in \Algnameabbr{} by comparing it with the baseline method that learns MI-based attribution. The baseline follows existing work~\cite{ying2019gnnexplainer, luo2020parameterized} to adopt the mutual information for the supervision of training the explainer.
Specifically, the parameters of amortized explainer are updated to minimize the loss function given by $L = \mathbb{E}_{v_i \sim \mathcal{V}, v_j \sim \mathcal{N}(v_i)} [\hat{\phi}_{j \to i} - I(v_j; \hat{y})]^2$, where the MI value is given by $I(v_j; \hat{y}) = - \mathbb{E}_{y \sim f(\mathcal{G}_{\mathcal{N}(v_i)}), \mathcal{S}_j \sim \mathcal{N}(v_i)} \log \frac{p(\hat{y} = y | \mathcal{G}_{\mathcal{N}(v_i)})}{ p(\hat{y} = y | \mathcal{G}_{\mathcal{S}_j})}$.
The experiment is conducted on the ogbn-arxiv dataset~(a real-world large graph), and the implementation details can be found in~\Cref{{supp:experiments}}.
Figure~\ref{fig:removal_vs_mi} shows the interpretation fidelity of \Algnameabbr{} compared to the baseline method, where \Algnameabbr{} consistently outperforms the baseline method. This is consistent with our observation in Section~\ref{sec:removal-based-attribution}, highlighting the advantage of our proposed removal-based attribution in generating faithful GNN explanations and serving as the objective function for training the amortized explainer. 

\begin{table}[t]
\begin{minipage}{.33\linewidth}
\centering
\includegraphics[width=.9\linewidth]{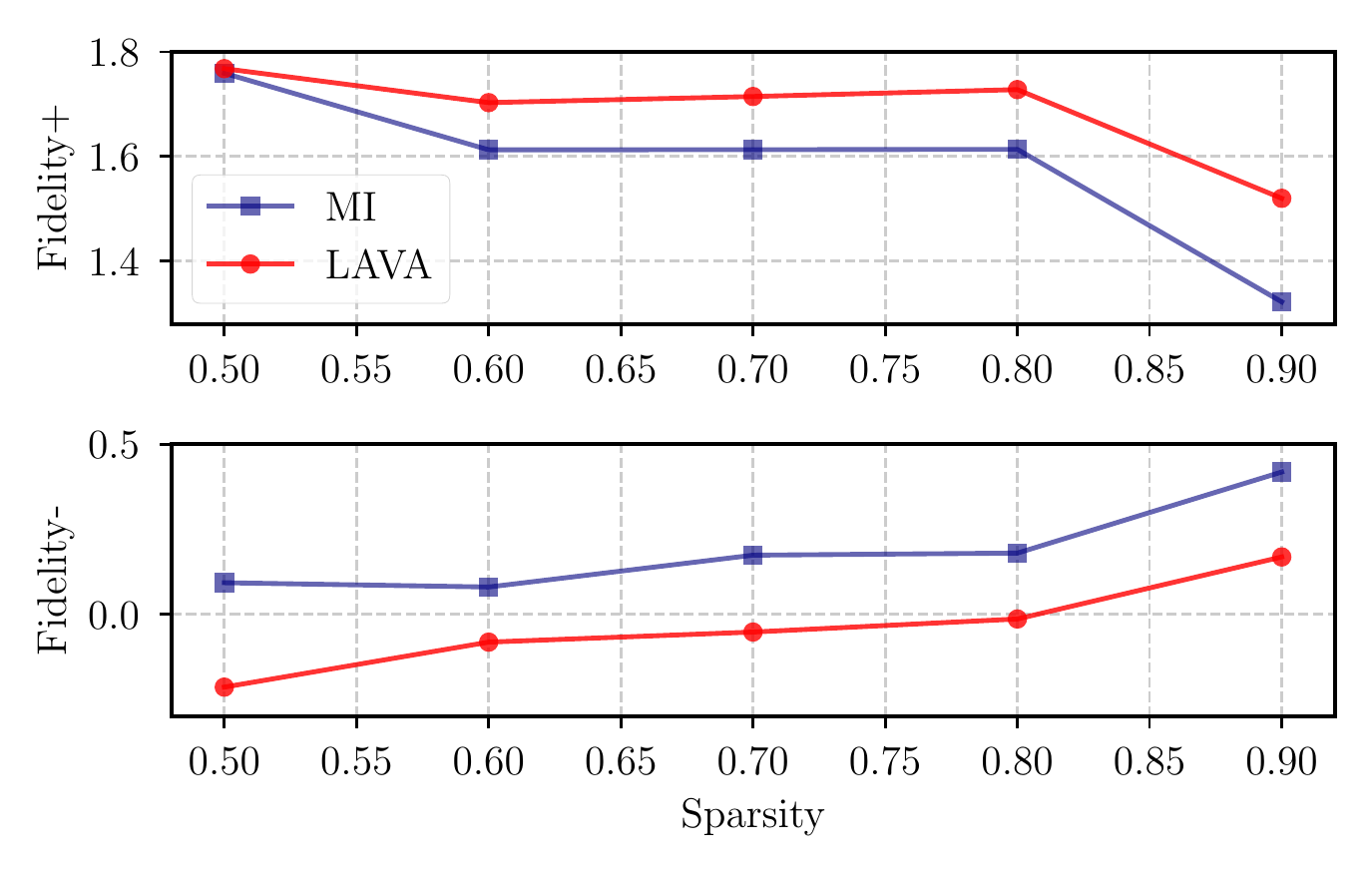}
\captionof{figure}{\Algnameabbr{} v.s. learning MI.}
\label{fig:removal_vs_mi}
\end{minipage}
\hfill
\begin{minipage}{.62\linewidth}
    \setlength{\abovecaptionskip}{0.5mm}
    \setlength{\belowcaptionskip}{-2mm}
    \centering
    \caption{\Algnameabbr{} v.s. single-embedding baseline method.}
    \label{tab:bidirection-attr}
    \centering
    \resizebox{1.\linewidth}{!}{
    \begin{tabular}{ll|cccc}
    \toprule[1pt]
      & Method & Tree-Cycles &	ogbn-arxiv	& BA-2Motif & MUTAG \\
    \midrule
      \multirow{2}{*}{$\mathrm{Fidelity}^+ \uparrow$} & Single & 1.61	& 1.18  &	0.60  & 1.31  \\
      & \Algnameabbr{} & \textbf{2.19}	& \textbf{1.54} &	\textbf{0.65} & \textbf{1.36}  \\
    \midrule
      \multirow{2}{*}{$\mathrm{Fidelity}^- \downarrow$} & Single & 1.23 & 0.16	& 0.58 & 1.24 \\
      & \Algnameabbr{} & \textbf{1.02} & \textbf{0.02}	& \textbf{0.55} & \textbf{1.20} \\
    \bottomrule[1pt]
    \end{tabular}
    }
\end{minipage}
\end{table}

\begin{table}[t]
    \begin{minipage}{.44\linewidth}
    \caption{Time latency ($s$) on ogbn-arxiv.}
    \label{tab:time_latency_arxiv}
    \centering
    \resizebox{\linewidth}{!}{
    \begin{tabular}{l|cccc}
    \toprule[1pt]
      Node \# & $10^2$ & $10^3$ & $10^4$ & $10^5$ \\
    \midrule
      GNNExplainer & 15.34 & 119.84  &  1186.43 & 12101.1 \\
      PGExplainer & 5.63 & 35.52  & 490.48  & 3534.65\\
      \Algnameabbr{} & \textbf{0.54} & \textbf{4.23} & \textbf{39.92}  & \textbf{416.50} \\
    \bottomrule[1pt]
    \end{tabular}
    }
    \end{minipage} 
    \hfill
    \begin{minipage}{.51\linewidth}
        \caption{Effectiveness of explainer layer number.}
        \label{tab:layer num}
        \centering
        \resizebox{.99\linewidth}{!}{
        \begin{tabular}{c|ccccc}
        \toprule[1pt]
        Aggr. Layer  & $0$ & $1$ & $2$ & $3$ & $4$\\\midrule
        $\mathrm{Fidelity}^+$ $\uparrow$  & $3.46$   & $3.25$   & $3.68$   & $\textbf{3.72}$  & $3.53$ \\ 
        $\mathrm{Fidelity}^-$  $\downarrow$ &  $2.34$  &  $2.52$  & $2.52$   & $\textbf{2.32}$ & $2.39$  \\
        Throughput & $\textbf{3308.77}$ & $1530.98$ & $903.30$ & $645.93$ & $499.71$\\
        \bottomrule[1pt]
        \end{tabular}
        }
    \end{minipage} 
\end{table}

\vspace{-6pt}
\paragraph{Effectiveness of Bidirectional Embedding.}
We demonstrate the advantage of \Algnameabbr{} by learning the source and target embedding separately in Eqauation~(\ref{eq:phi_hat}). 
In contrast, a baseline method is to learn a single embedding $\mathbf{t}$ by the explainer to estimate the node attribution.
Concretely, the attribution of a node $v_j$ to another node $v_i$ is given by $\hat{\phi}_{j \to i} = \langle \mathbf{t}_j, \mathbf{t}_i \rangle$. In this case, $\hat{\phi}_{i \to j} = \hat{\phi}_{j \to i}$, and its attribution to a graph $\mathcal{G}$ is estimated based on the pooling of $\mathbf{t}_i$ for $v_i \in \mathcal{G}$.
\Cref{tab:bidirection-attr} compares the fidelity of \Algnameabbr{} versus the baseline on both node classification and graph classification tasks: the Tree-Cycles, ogbn-arxiv, BA-2Motif, and MUTAG datasets. It is observed that \Algnameabbr{} consistently outperforms the baseline. In particular, in node classification, \Algnameabbr{} shows a $0.58$  improvement in $\mathrm{Fidelity^+}$, and a $0.15$ reduction in $\mathrm{Fidelity^-}$. This emphasizes the advantage and necessity of \Algnameabbr{} in learning the source and target embedding separately for each node to achieve faithful explanations.

\vspace{-5pt}
\paragraph{Explainer with Different Numbers of Layers.}
We study the performance of \Algnameabbr{} affected by the number of aggregation layers in the amortized explainer $\textsl{g}(\bullet~|~ \theta_{\textsl{g}})$.
\Cref{tab:layer num} lists the fidelity of \Algnameabbr{} with respect to different numbers of aggregation layers in the amortized explainer, where the explainer without aggregation is essentially a multi-layer perceptron~(MLP) model.
It is observed that the GNN-based explainer~(\#AggrLayer = 3) achieves the highest fidelity. The performance on fidelity metrics increases with the number of aggregation layers, suggesting that the node information aggregation in the framework of \Algnameabbr{} benefits the training of amortized explainer. Nevertheless, too many layers (\#AggrLayer $>$ 3) could result in redundancy and thus hurt the performance.
Moreover, an inspirative observation is that \Algnameabbr{} has moderate performance in fidelity without node information aggregation, where $\mathrm{Fidelity}^+ \!=\! 3.46$ and $\mathrm{Fidelity}^- \!=\! 2.34$.
This indicates the effectiveness of \Algnameabbr{} even without a GNN-based explainer.
Notably, since an MLP-based explainer enables a significant improvement in throughput, implementing \Algnameabbr{} with an MLP-based explainer highlights a potential solution for industrial applications.

\begin{figure*}[t]
\setlength{\belowcaptionskip}{-1mm}
\centering
\begin{subfigure}[b]{0.1\textwidth}
\centering
        \includegraphics[width=1.0\linewidth]{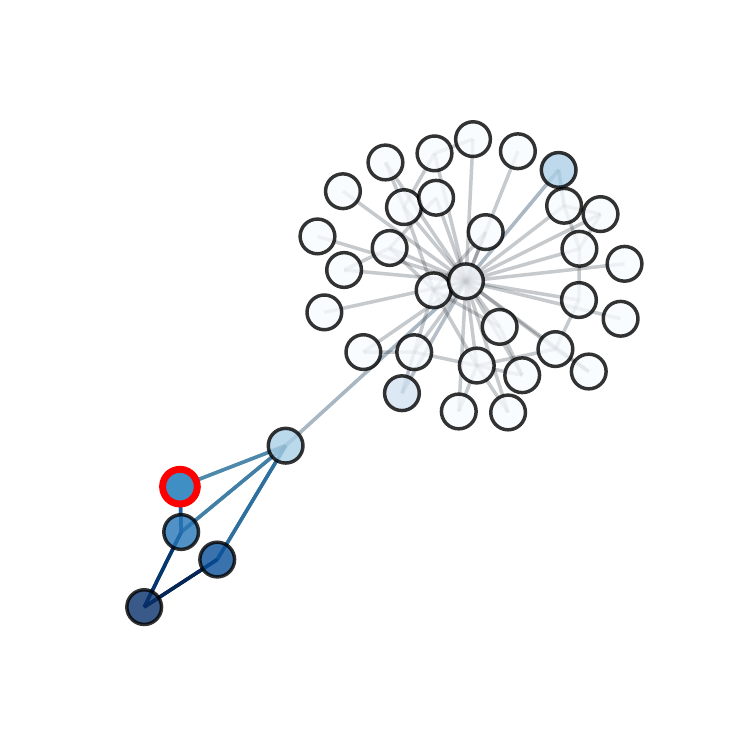}
\end{subfigure}%
\hfill
\begin{subfigure}[b]{0.1\textwidth}
\centering
        \includegraphics[width=1.0\linewidth]{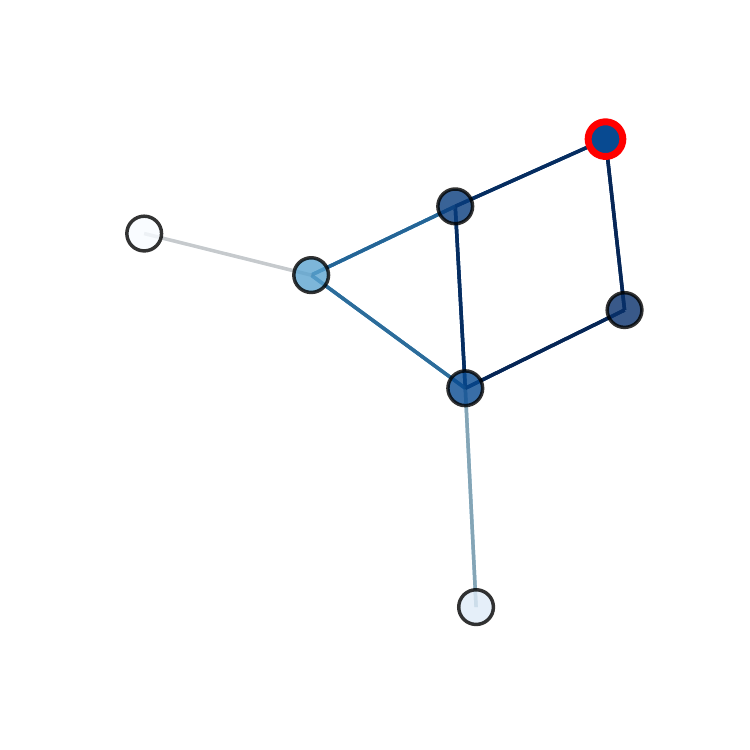}
\end{subfigure}%
\hfill
\begin{subfigure}[b]{0.1\textwidth}
\centering
        \includegraphics[width=1.0\linewidth]{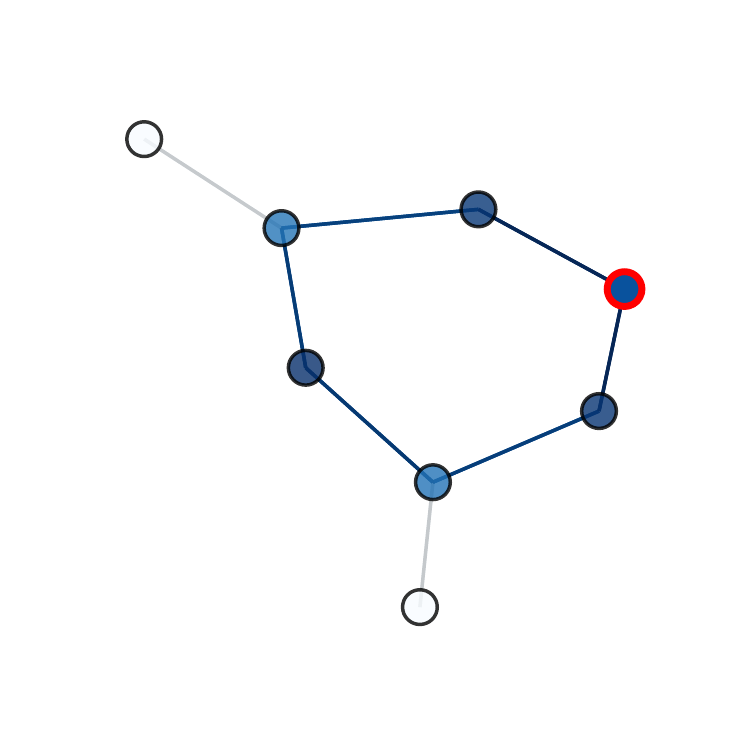}
\end{subfigure}%
\hfill
\begin{subfigure}[b]{0.1\textwidth}
\centering
        \includegraphics[width=1.0\linewidth]{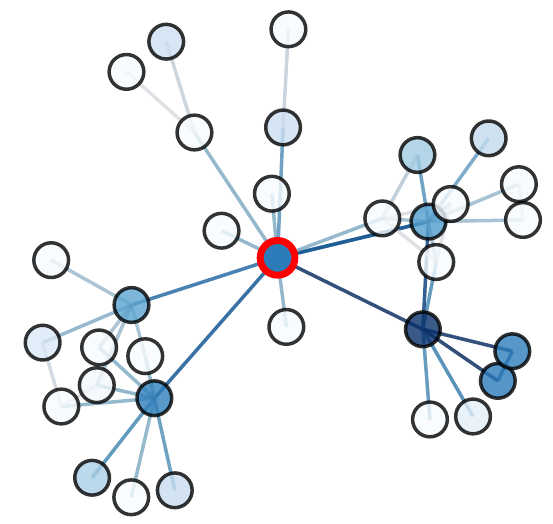}
\end{subfigure}
\hfill
\begin{subfigure}[b]{0.1\textwidth}
\centering
        \includegraphics[width=1.0\linewidth]{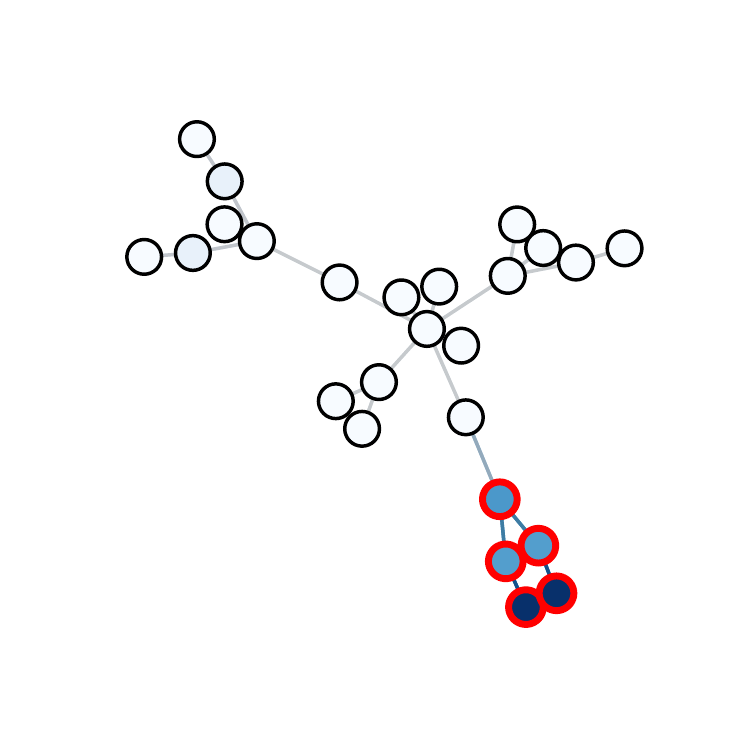}
\end{subfigure}%
\hfill
\begin{subfigure}[b]{0.1\textwidth}
\centering
        \includegraphics[width=1.0\linewidth]{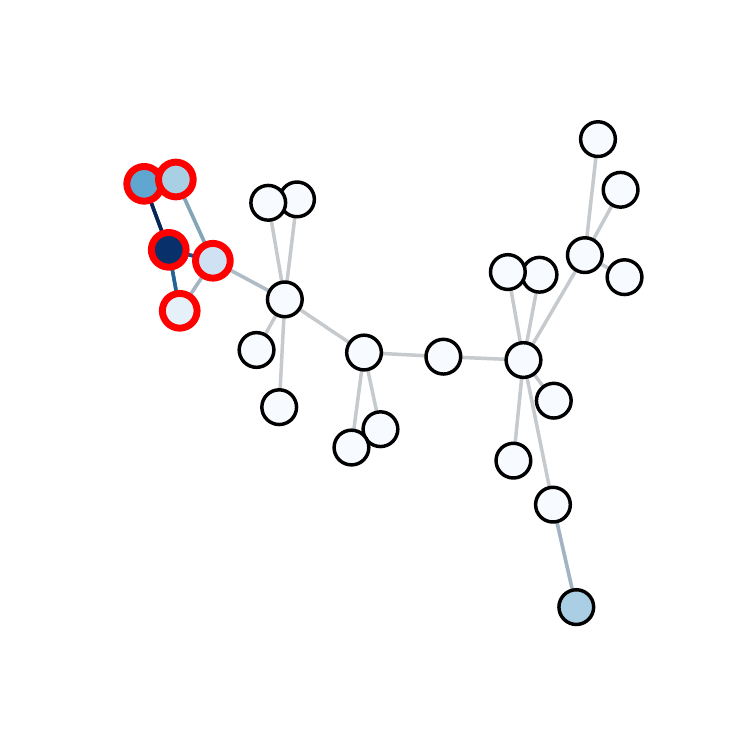}
\end{subfigure}%
\hfill
\begin{subfigure}[b]{0.1\textwidth}
\centering
        \includegraphics[width=1.0\linewidth]{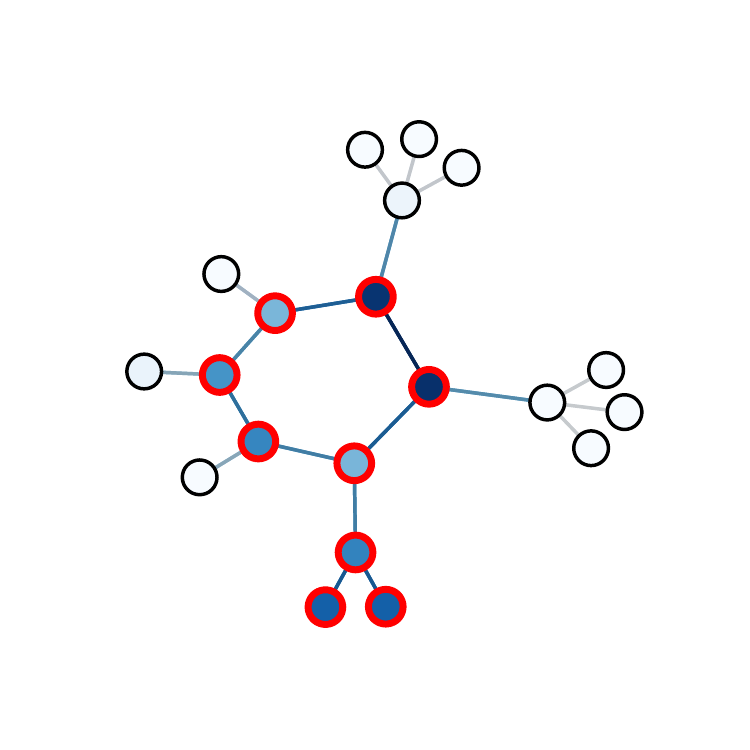}
\end{subfigure}%
\\
\begin{subfigure}[b]{0.1\textwidth}
\centering
        \includegraphics[width=1.0\linewidth]{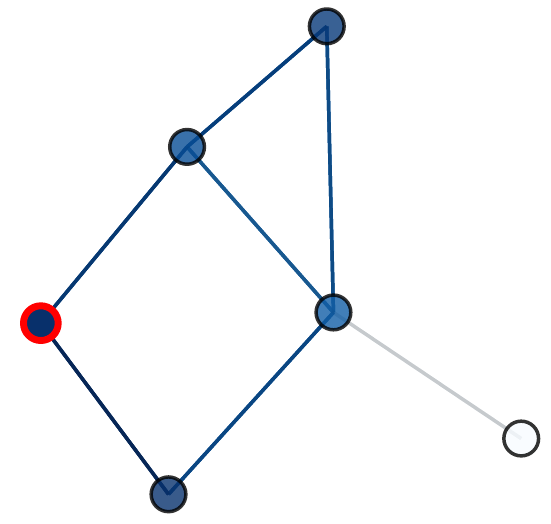}
        \caption{}
        \label{subfig:bashapes1}
\end{subfigure}
\hfill
\begin{subfigure}[b]{0.1\textwidth}
\centering
        \includegraphics[width=1.0\linewidth]{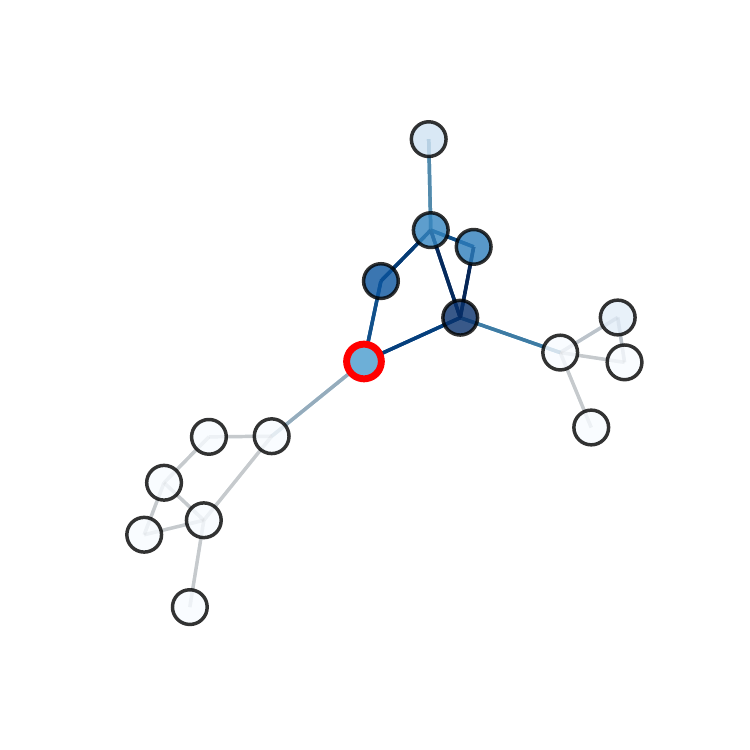}
        \caption{}
        \label{subfig:bacomm1}
\end{subfigure}%
\hfill
\begin{subfigure}[b]{0.1\textwidth}
\centering
        \includegraphics[width=1.0\linewidth]{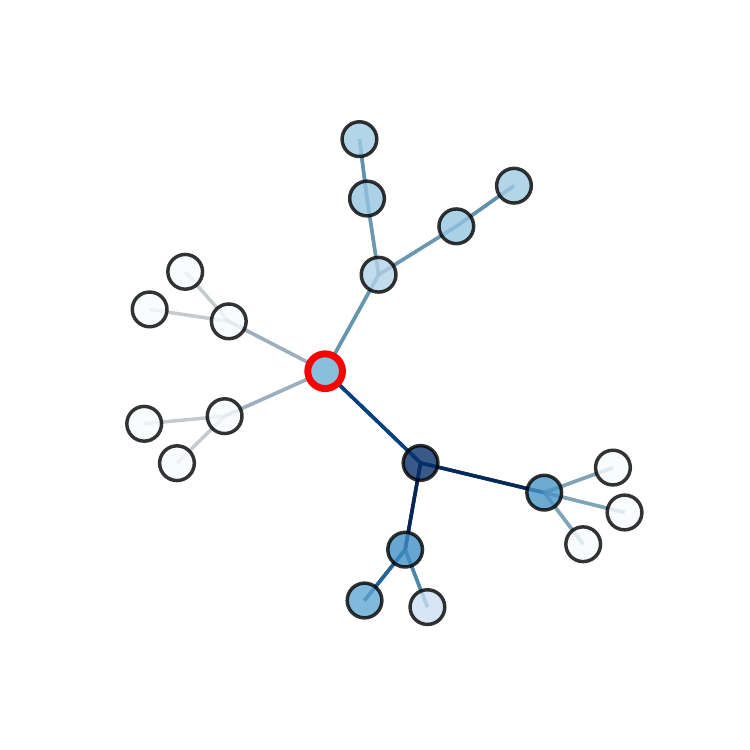}
        \caption{}
        \label{subfig:tree1}
\end{subfigure}%
\hfill
\begin{subfigure}[b]{0.1\textwidth}
\centering
        \includegraphics[width=1.0\linewidth]{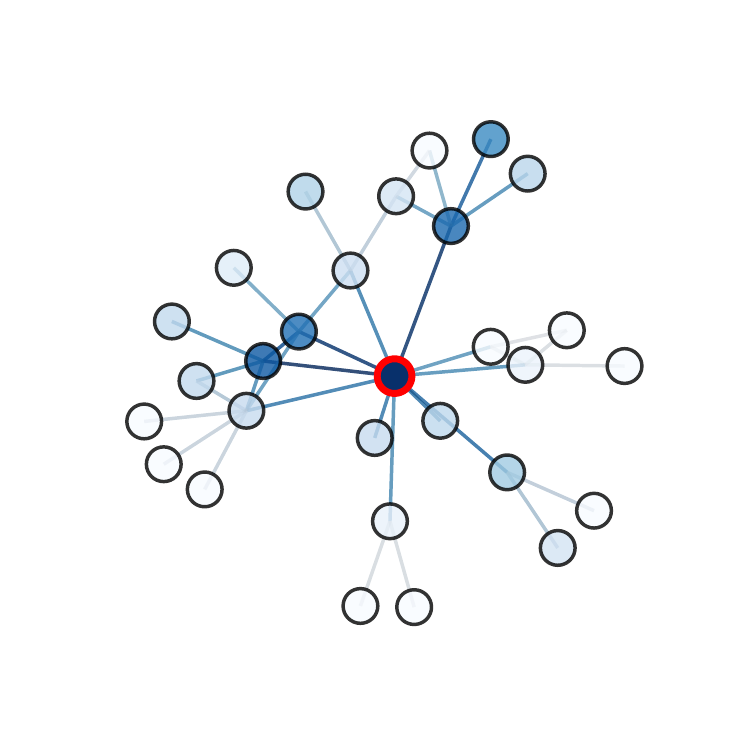}
        \caption{}
        \label{subfig:arxiv1}
\end{subfigure}%
\hfill
\begin{subfigure}[b]{0.1\textwidth}
\centering
        \includegraphics[width=1.0\linewidth]{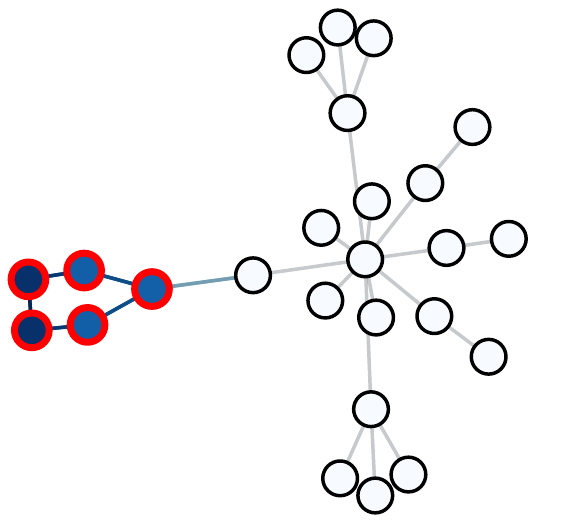}
        \caption{}
        \label{subfig:ba2-0-1}
\end{subfigure}
\hfill
\begin{subfigure}[b]{0.1\textwidth}
\centering
        \includegraphics[width=1.0\linewidth]{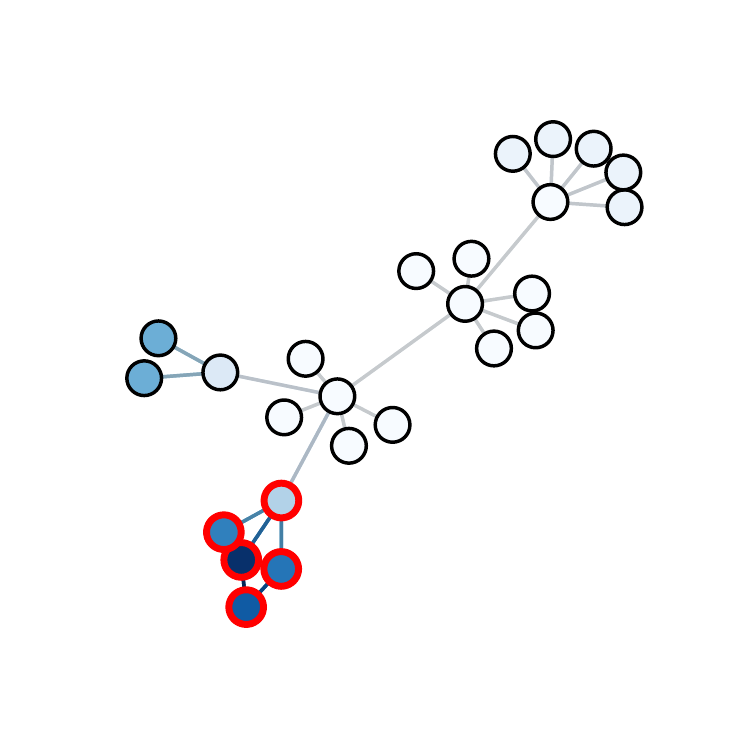}
        \caption{}
        \label{subfig:ba2-1-1}
\end{subfigure}%
\hfill
\begin{subfigure}[b]{0.1\textwidth}
\centering
        \includegraphics[width=1.0\linewidth]{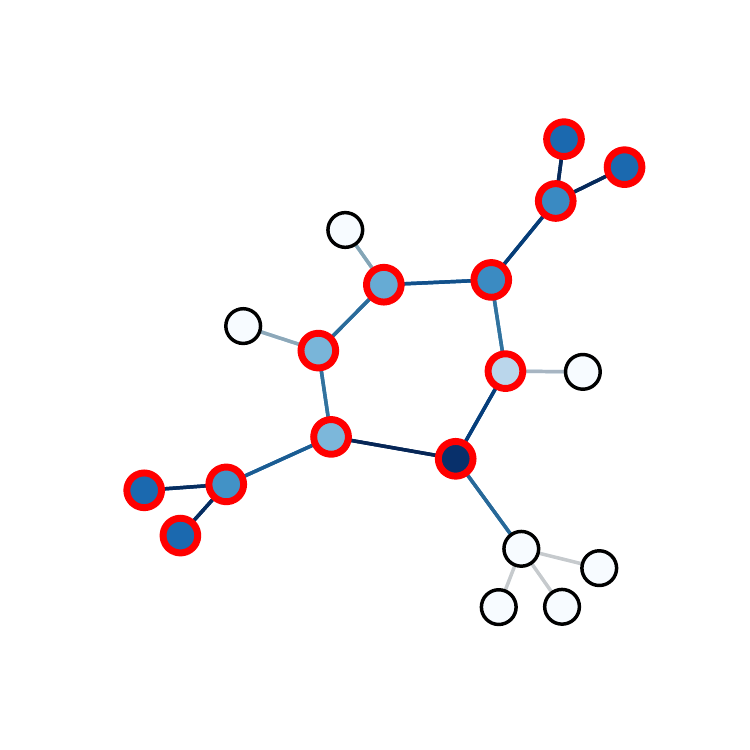}
        \caption{}
        \label{subfig:mutag-0-1}
\end{subfigure}%
\caption{Qualitative results of node and graph classification. \textbf{Node classification}: (a) BA-Shapes (b) BA-Community (c) Tree-Cycles (d) ogbn-arxiv; \textbf{Graph classification} (e) BA-2Motifs, Pentagon (f) BA-2Motifs, House (g) MUTAG, Mutagen.
For node classification, darker nodes refer to more important nodes for the prediction of the target node (to be explained and highlighted in red color) in~(a)(b)(c)(d). 
For graph classification, darker nodes indicate more important nodes for the classification, and red nodes indicate the ground truth motif in~(e)(f)(g). 
}
\label{fig:quali node cls}
\vspace{-7pt}
\end{figure*}

\vspace{-8pt}
\subsection{Qualitative Results}
\label{sec:quali res}

In this section, we visualize the explanations generated by \Algnameabbr{}, demonstrating its power in helping human users understand GNN models. 
The explanation of node classification is shown in \Cref{fig:quali node cls}~(a)-(d), where the target node is highlighted in red, and the dark blue nodes indicate important neighbors suggested by \Algnameabbr{}. The explanation of graph classification is shown in \Cref{fig:quali node cls}~(e)-(g), where the ground truth explanation is marked in red, and the recognized important nodes are in dark blue. 

\vspace{-7pt}
\paragraph{Node Classification.}
The classification task aims to predict whether a target node belongs to a specific structural subgraph on the BA-shapes and Tree-Cycles datasets, or a community on the BA-Community dataset.
According to the explanation given by \Algnameabbr{} in \Cref{fig:quali node cls}~(a)-(d), the important neighbors for the target nodes fall into a local subgraph with a specific structure. e.g. a motif of ``house'' on the BA-shapes and BA-Community datasets, a hexagon or tree structures on Tree-Cycles. Such faithful and intuitive explanations are essential to increase user trust in practical use cases.
Different phenomena can be observed on the ogbn-arxiv dataset as demonstrated in \Cref{fig:quali node cls}~(d), where no ground truth pattern is available.
Concretely, the direct neighbors have more contribution in the prediction of the target node than the non-adjacent nodes.
On the ogbn-arxiv dataset, each node represents a paper, each edge indicates a citation of another one, and the classification aims to predict whether a paper belongs to one of the 40 arXiv CS subject areas. As common sense, the papers cited with each other likely belong to the same subject area. 
Explanations in \Cref{fig:quali node cls}~(d) reflect the intuition of this task, indicating the effectiveness of \Algnameabbr{}.


\vspace{-7pt}
\paragraph{Graph Classification.} 
The graph classification aims to detect whether a graph has a specific subgraph. According to the ground truth explanation, the key subgraph structure for the classification is a pentagon or a house structure on the BA-2Motifs dataset. Notably, the nodes on the edge inside the house are the most important ones as highlighted in \Cref{fig:quali node cls}~(f).
On MUTAG, the ground truth for the class mutagen is a hexagon (known as a ``carbon ring'') with a three-node (NH$_2$/NO$_2$) group, which is recognized by \Algnameabbr{} in \Cref{fig:quali node cls}~(g).
As shown in \Cref{fig:quali node cls}~(e)(f)(g), \Algnameabbr{} successfully generates explanations of each graph consistent with the ground truth.
The results validate the capability of \Algnameabbr{} in explaining graph classification. 

%% file: text/7-conclusion.tex
\vspace{-8pt}
\section{Conclusion}
In this work, we propose \Algnameabbr{} for efficient and faithful GNN explanations. Specifically, we introduce a removal-based attribution approach, which outperforms the mutual information as guidance of generating faithful explanations, especially on large graphs. The correlation between removal-based attribution and interpretation fidelity is theoretically and experimentally validated. Following this direction, \Algnameabbr{} trains an amortized explainer to maximize fidelity and incorporates a subgraph sampling strategy for scalability on large graphs. After training, \Algnameabbr{} generates batch-wise node explanations efficiently in a feed-forward process. Experimental results demonstrate \Algnameabbr{} can significantly improve the efficiency and scalability of GNN explanations, while keeping competitive interpretation fidelity.
This highlights the potential of \Algnameabbr{} being applied to the large-scale social graphs with millions of nodes and edges in real-world scenarios.



%% file: text/8-appendix.tex
\newpage
\appendix
\onecolumn

\section{Experimental Details in Section~3}
\label{supp:removal-based attribution experiment}
In this section, we provide implementation details of the experiment in Section~3. In order to validate the correlation between MI- or removal-based attribution value and the fidelity score, we randomly sample a subset of neighbor nodes $\mathcal{J} \subseteq \mathcal{N}(v_i)$ of a target node $v_i$ as the explanatory nodes (important nodes to the model prediction). Then, we calculate the MI- or removal-based attribution value of $\mathcal{J}$ to the original prediction for the target node $f(\mathcal{G}_{\mathcal{N}(v_i)})$ following previous work~\cite{ying2019gnnexplainer}. Similar to Definition~3.2, we denote the MI value of a subset $\mathcal{J}$ to a target node $v_i$  as $\alpha_{\mathcal{J} \to i}$. Likewise, removal-based attribution of the subset $\mathcal{J}$ to $v_i$ is denoted as $\phi_{\mathcal{J} \to i}$. 
Concretely, we calculate the MI value $\alpha_{\mathcal{J} \to i}$ as follows:
\begin{equation}
    \alpha_{\mathcal{J} \to i} = - \mathbb{E}_{y \sim f(\mathcal{G}_{\mathcal{N}(v_i)})} p(\hat{y} = y | \mathcal{G}_{\mathcal{N}(v_i)}) \log p(\hat{y} = y | \mathcal{G}_{\mathcal{J}}).
\end{equation}
For the same $\mathcal{J}$, we calculate the removal-based attribution score as:
\begin{equation}
    \phi_{\mathcal{J} \to i} = \mathbb{E}_{j \sim \mathcal{J}}\big[ f(\mathcal{G}_{v_j}) - f(\mathcal{G}_{\mathcal{N}({v_i)} \setminus v_j }) \big].
\end{equation}
For each subfigure in Figure~2 in Section~3, we randomly sample one hundred input pairs $(\mathcal{J}, v_i)$ to calculate the MI or removal-based attribution values, which are plotted on the x-axis. 

Given $(\mathcal{J}, v_i)$, $\mathrm{Fidelity^+}$ is calculated as follows:
\begin{equation}
    \mathrm{Fidelity}^+(v_i, \mathcal{J}) = \mathcal{G}_{\mathcal{N}(v_i)} - f( \mathcal{G}_{\mathcal{N} (v_i) \setminus \mathcal{J}}),
\end{equation}
which is plotted on the y-axis.
For Figure~2~(a), we use the dataset BA-Community to conduct the experiment, while ogbn-arxiv is used in Figure~2~(b) and Figure~2~(c).

\section{Proof of Theorem 3.1}
\label{supp:theorem proof}
In this section, we provide the proof of Theorem~3.1.

\begin{proof}
    We take $\phi_{j \to i} =  \mathbb{E}_{\mathcal{S}_j \subseteq \mathcal{N}(v_i)} \big[ f(\mathcal{G}_{\mathcal{S}_j}) - f(\mathcal{G}_{\mathcal{N}({v_i)} \setminus \mathcal{S}_j }) \big]$ into  $\mathbb{E}_{j \sim \mathcal{J}} \phi_{j \to i}$,
    \begin{align}
        & \mathbb{E}_{j \sim \mathcal{J}}\mathbb{E}_{\mathcal{S}_j \subseteq \mathcal{N}(v_i)} \big[ f(\mathcal{G}_{\mathcal{S}_j}) - f(\mathcal{G}_{\mathcal{N}({v_i)} \setminus \mathcal{S}_j }) \big] \\
        =&\mathbb{E}_{j \sim \mathcal{J}}\mathbb{E}_{\mathcal{S}_j \subseteq \mathcal{N}(v_i)} \big[ \underbrace{(f(\mathcal{G}_{\mathcal{N}(v_i)}) - f(\mathcal{G}_{\mathcal{N}({v_i)} \setminus \mathcal{S}_j }))}_{\mathrm{Fidelity}^+(v_i, \mathcal{S}_j)} -\underbrace{(f(\mathcal{G}_{\mathcal{N}(v_i)})-f(\mathcal{G}_{\mathcal{S}_j}))}_{\mathrm{Fidelity}^-(v_i, \mathcal{S}_j)}\big]    \\
        =& \mathbb{E}_{j \sim \mathcal{J}}\mathbb{E}_{\mathcal{S}_j \subseteq \mathcal{N}(v_i)} [\Delta \mathrm{Fidelity}(v_i, \mathcal{S}_j)].
    \end{align}
    Likewise, we get 
    \begin{align}
        & \mathbb{E}_{j \sim \mathcal{J^*}}\mathbb{E}_{\mathcal{S}_j \subseteq \mathcal{N}(v_i)} \big[ f(\mathcal{G}_{\mathcal{S}_j}) - f(\mathcal{G}_{\mathcal{N}({v_i)} \setminus \mathcal{S}_j }) \big] \\
        =& \mathbb{E}_{j \sim \mathcal{J^*}}\mathbb{E}_{\mathcal{S}_j \subseteq \mathcal{N}(v_i)} [\Delta \mathrm{Fidelity}(v_i, \mathcal{S}_j)].
    \end{align}
    Since $\mathbb{E}_{j \sim \mathcal{J}} \phi_{j \to i} \leq \mathbb{E}_{j \sim \mathcal{J}^*} \phi_{j \to i}$, we get
     \begin{align}
        \mathbb{E}_{j \sim \mathcal{J}} \mathbb{E}_{\mathcal{S}_j \subseteq \mathcal{N}(v_i)} \big[ f(\mathcal{G}_{\mathcal{S}_j}) - f(\mathcal{G}_{\mathcal{N}({v_i)} \setminus \mathcal{S}_j }) \big]  &\leq \mathbb{E}_{j \sim \mathcal{J}^*} \mathbb{E}_{\mathcal{S}_j \subseteq \mathcal{N}(v_i)} \big[ f(\mathcal{G}_{\mathcal{S}_j}) - f(\mathcal{G}_{\mathcal{N}({v_i)} \setminus \mathcal{S}_j }) \big] \\
    \mathbb{E}_{j \sim \mathcal{J}} \mathbb{E}_{\mathcal{S}_j \subseteq \mathcal{N}(v_i)}[\Delta \mathrm{Fidelity}(v_i, \mathcal{S}_j)] &\leq \mathbb{E}_{j \sim \mathcal{J}^*} \mathbb{E}_{\mathcal{S}_j \subseteq \mathcal{N}(v_i)}[\Delta \mathrm{Fidelity}(v_i, \mathcal{S}_j)].
    \end{align}

\end{proof}

\section{Experimental Details of Figure~1}
\label{supp:details_radar}
We plot $\mathrm{Fildelty^+}$ and throughput in the radar figure for four baselines and our method. $\mathrm{Fildelty^+}$ is the averaged value over different sparsity given in Figure~4 (subgraphs of $\mathrm{Fidelity^+}$) in Section~5.1. The throughput in the plot is given in Table~1 in Section~5.1.

\section{Experimental Details in Section~5}
\label{supp:experiments}
In this section, we introduce details about the used datasets, target models, baseline models, hyper-parameter settings, implementation details of experiments in Section~5.2, and used computational infrastructure for our experiments. 
\subsection{Datasets}
We deploy \Algnameabbr{} to explain node classification on BA-Shapes, BA-Community, Tree-Cycles and ogbn-arxiv; and graph classification on the BA-2Motifs and MUTAG datasets. 
In particular, the BA-Shapes dataset contains a 300-node Barab\'asi-Albert (BA) graph \cite{barabasi1999emergence} with 80 five-node ``house'' motifs randomly attached to it. Each node has a label out of four according to the position in the motif. 
The BA-Community dataset consists of two BA-Shapes datasets with random edges connecting them, including eight classes. 
The Tree-Cycles dataset has two node classes indicating whether a node belongs to a base graph (eight-level balanced binary tree) or a motif (six-node cycle). 
The BA-2Motifs has $1,000$ synthetic graphs with binary labels, while MUTAG contains $4,337$ molecule graphs with binary labels.
The statistics of each dataset can be found in \Cref{tab:dataset stats}. For all datasets, the split of training/validation/test set is 80\%/10\%/10\%.

\begin{table}[t]
\centering
\caption{Dataset details. The upper part contains the node classification task and the bottom part is graph classification.}
\label{tab:dataset stats}
\begin{tabular}{cccccc}
\toprule[1pt]
             & \#graphs & \#nodes & \#edges & \#labels & node feature dim \\\hline
BA-Shapes    &     1     &    700     &    4,110     &   4  &  10                \\
BA-Community &    1      &    1,400   &   8,920      &   8  &  10                \\
Tree-Cycles  &     1     &    871     &   1,950      &  2  &   10               \\
ogbn-arxiv   &     1     &    169,343     &  1,166,243    &  40  &   128            \\
\midrule[1pt]
BA-2Motifs   &   1,000       &  25,000     &  51,392   &  2   &   10              \\
MUTAG        &    4,377      &   131,488      & 266,894   &  2  & 14  \\          \bottomrule[1pt]
\end{tabular}
\end{table}

\begin{figure}[h]
    \centering
    \includegraphics[width=.9\linewidth]{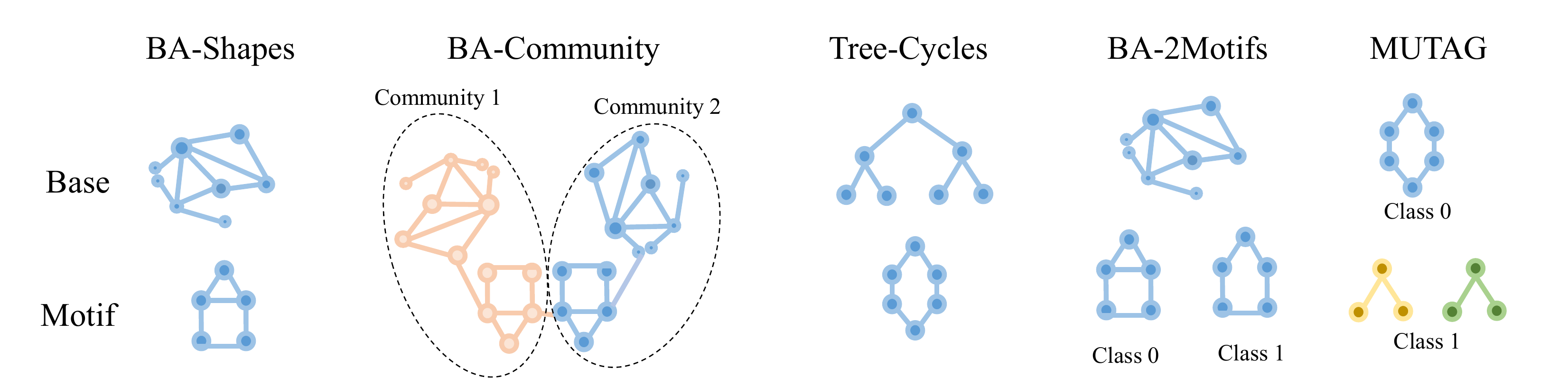}
    \caption{Ground truth motif on BA-Shapes, BA-Community, Tree-Cycles (node classification datasets), BA-2Motif and MUTAG (graph classification datasets).}
    \label{fig:gt motif}
\end{figure}

Ground truth motifs on synthesis datasets and MUTAG are listed in \Cref{fig:gt motif}. On BA-Shapes and BA-Community, the labels of nodes are decided by the position of the node in the ``house'' motif, while on Tree-Cycles the label of nodes is dependent whether the node is inside the six-node cycle or on the tree structure. On BA-2Motifs, the graph is classified by the type of motifs. In MUTAG, there are two different classes: mutagen and non-mutagen. If there is a pattern of NH$_2$/NO$_2$ together with the carbon ring, then this graph belongs to the class mutagen, otherwise to the class Non-mutagen.

To the best of our knowledge, this is the first work to benchmark the explanations on large graph datasets like ogbn-arxiv~\cite{hu2020open}. 
The ogbn-arxiv dataset is essentially a citation network, containing a total of $169,343$ nodes and $1,166,243$ edges, where a node represents a paper and an edge indicates a citation of one paper by another one. 
Each node is encoded in a 128-dim feature vector provided by the Word2Vec model with the MAG corpus~\cite{mikolov2013distributed}. 
On this dataset, the target GNN model is trained to classify a paper into one of the 40 arXiv CS subject areas.


\subsection{Target Model Details}
For each dataset, we use the target model under the settings as given in \cite{luo2020parameterized} for BA-Shapes, BA-Community, Tree-Cycles, BA-2Motifs and MUTAG~\footnote{\tiny We use the trained target models at \url{https://github.com/LarsHoldijk/RE-ParameterizedExplainerForGraphNeuralNetworks}.}. For each of these datasets, a GCN model containing three layers is trained. 
In the graph classification models, a max-pooling layer is inserted to get the graph embedding before the final fully-connected layer.
On ogbn-arxiv, we train a GraphSAINT~\cite{zeng2019graphsaint} model with two layers using a subgraph-wise sampling strategy following the settings in \cite{duan2022comprehensive}\footnote{\tiny Code for training GraphSAINT on ogbn-arxiv: \url{https://github.com/VITA-Group/Large_Scale_GCN_Benchmarking}.}. 
For the synthetic datasets, we use the same splits provided in \cite{luo2020parameterized}. For MUTAG and ogbn-arxiv, we use their original splits. On all datasets, we use the identical target model as well as the same data split for GNNExplainer, PGExpaliner, GraphSVX, OrphicX and \Algnameabbr{} to conduct a fair comparison of their interpretation.

\Cref{tab:target model} lists the accuracy of the target models on different splits on each dataset.


\begin{table}[h]
\centering
\resizebox{\linewidth}{!}{
\begin{tabular}{ccccccc}
\toprule[1pt]
           & BA-Shapes & BA-Community & Tree-Cycles & ogbn-arxiv & BA-2Motifs & \multicolumn{1}{c}{MUTAG} \\\hline
Training   &   0.97 &    0.90          &    0.94         &    0.56        &     1.0       &0.82 \\
Validation &     1.00      & 0.76 &     0.98        &      0.56     & 1.0           &      0.82                     \\
Test       &    1.00       &    0.72          &     0.94        &        0.50    &     1.00       &  0.81   \\
\bottomrule[1pt]
\end{tabular}
}
\caption{Accuracy of target models.}
\label{tab:target model}
\end{table}


\subsection{Baseline Methods} 
We introduce in this section the used state-of-the-art baseline methods in comparison to our proposed framework \Algnameabbr{}. \textbf{GNNExplainer} \cite{ying2019gnnexplainer} explains predictions via soft masks for edges and node features, which maximizes the mutual information between the predictions of original graphs and the newly generated ones. For each instance, a soft mask needs to be learned. \textbf{GraphSVX}~\cite{duval2021graphsvx} also provides local explanations. The authors propose to estimate Shapley values for nodes (and features on a node) adopted kernel SHAP~\cite{lundberg2017unified} and use SHAP values as model explanations. 
\textbf{PGExplainer} \cite{luo2020parameterized} learns a model using the reparameterization trick to predict edge masks indicating their importance, while \textbf{OrphicX}~\cite{lin2022orphicx} identifies the causal factors by maximizing the information flow from the latent features to the model predictions, which are used to produce explanations. All methods estimate the explanation of a node by averaging the importance scores of the edges connected to the node, if the original method is designed for explaining through edge importance. We adopt code and follow their hyper-parameter settings for training the explainer given in the official code for baseline methods. For GraphSVX, we deploy the uniform mask sampling for estimating SHAP values~\footnote{\tiny Code for GNNExplainer and PGExplainer is at \url{https://github.com/LarsHoldijk/RE-ParameterizedExplainerForGraphNeuralNetworks}; for OrphicX is at \url{https://github.com/wanyugroup/cvpr2022-orphicx}; for GraphSVX is at \url{https://github.com/AlexDuvalinho/GraphSVX}.}.

\subsection{Hyper-parameter Settings} 
We adopt a GCN-based model to build the explainer $\textsl{g}(\bullet ~|~ \theta_{\textsl{g}}^*)$.
The number of aggregation layers in the explainer is set to the Max Hop value defined in Section~4.2. For the explanation of graph classification, the target embedding $\mathbf{t}_{\mathcal{G}}$ takes the max-pooling of the target embeddings $\mathbf{t}_j$ for nodes $v_j \in \mathcal{G}$. 
The explainer is trained with the default splitting of the training/validation/testing sets on each dataset. 
To update the parameters of $\textsl{g}(\bullet ~|~ \theta_{\textsl{g}}^*)$, we deploy the Adam optimizer with a $10^{-3}$ learning rate for node classification and a $10^{-4}$ learning rate for graph classification.
The mini-batch size of sampling target nodes is set to $64$ for all datasets. 

\subsection{Fidelity Average}
Table~2 and~4 in Section~5.2 report the averaged fidelity values. We first calculate $\mathrm{Fidelity^+}$ and $\mathrm{Fidelity^-}$ at different sparsity (see Figure~4) and then use the averaged value to represent the final fidelity in these tables. We take a wide range of sparsity (from 0.3 to 0.7 with a step size of 0.1) in these two experiments. Following \cite{yuan2022explainability}, the sparsity of a given node $v_i$ is defined as:
\begin{equation}
    \mathrm{Sparsity} (v_i, \mathcal{N}^*_{v_i}) = 1 - \frac{|\mathcal{N}^*_{v_i}|}{|\mathcal{N}(v_i)|},
\end{equation}
where $\mathcal{N}^*_{v_i}$ is the set of important neighbor nodes given by the explanation. In $\mathrm{Fidelity^+}$, the node set $\mathcal{N}^*_{v_i}$ is removed, while in $\mathrm{Fidelity^-}$, $\mathcal{N}^*_{v_i}$ is kept but the rest nodes are removed. A higher sparsity is desired for an explanation since it captures sufficiently important information in a small subset of nodes. Through different sparsity, we control the number of the nodes inside $\mathcal{N}^*_{v_i}$ to fairly compare different methods.
\subsection{Computational Infrastructure Details}
\label{supp:infrastructure}
All experiments in this paper are conducted on the device given in \Cref{supp-tab:device}.
\begin{table}[h]
\centering
\caption{Computational infrastructure details.}
\label{supp-tab:device}
\begin{tabular}{c|c}
\toprule[1pt]
Device Attribute  & Value \\\hline 
Computing infrastructure&     GPU              \\
GPU model &    NVIDIA GeForce RTX 2080 Ti              \\
GPU number  &     1                   \\
CUDA version   &     11.3             \\
\bottomrule[1pt]
\end{tabular}
\end{table}

\section{Impact of Source and Target Embedding Dimensions}
We run experiments to study the performance of \Algnameabbr{} affected by the dimension of explanation-oriented embeddings $n$~(the dimension of $\mathbf{p}_i$ and $\mathbf{t}_i$ of each node $v_i \in \mathcal{V}$) in the explainer $\textsl{g}(\bullet~|~ \theta_{\textsl{g}})$. \Cref{tab:hidden dim} gives the results of $\mathrm{Fidelity^+}$ and $\mathrm{Fidelity^-}$ with $n \in \{10, 20, 30, 40\}$. Fidelity scores are averaged over the sparsity from 0.3 to 0.7 with a step size of 0.1.
It is observed that the optimal setting is $n \!=\! 20$ with $\mathrm{Fidelity^+}$ of $3.72$ and $\mathrm{Fidelity^-}$ of $2.32$.
This indicates a latent space of $n \!=\! 20$ provides an optimal solution for the hidden dimension of the explainer in \Algnameabbr{} to learn removal-based attribution.
In the case of $n \!\geq\! 20$, the redundancy of embedding causes overfit on the training dataset, which leads to a reduction of generalization fidelity. 
The experiments in this section are conducted on the BA-Community dataset under a unified setting of remaining factors. 

\begin{table}[h]
    \caption{Results using different dimensions of explanation-oriented embeddings.}
    \label{tab:hidden dim}
    \centering
    \resizebox{.5\linewidth}{!}{
    \begin{tabular}{c|cccc}
    \toprule[1pt]
    Dimension  & $10$ & $20$ & $30$ & $40$ \\\hline
    $\mathrm{Fidelity}^+$ $\uparrow$  & $3.44$   & $\textbf{3.72}$   & $3.42$   & $3.30$   \\ 
    $\mathrm{Fidelity}^-$ $\downarrow$ &  $2.53$  &  $\textbf{2.32}$  & $2.81$   & $2.82$   \\
    \bottomrule[1pt]
    \end{tabular}
    \vspace{-2pt}
    }
\end{table}

\section{Evaluation with Ground-truth}
On many synthetic datasets, ground-truth i.e., important nodes, are available. Therefore, many previous works such as \cite{ying2019gnnexplainer, luo2020parameterized, duval2021graphsvx} evaluate based on ground-truth. As given in the ground-truth, a node is labeled with $1$ if it is important for the decision while with $0$ if it is not. AUC is calculated between the given explanation scores and the ground-truth labels. We benchmark our methods with GNNExplainer, PGExplainer, and GraphSVX on synthetic datasets in \Cref{tab:auc}. Nodes/Graphs in the test set are used for AUC evaluation. 
We see that \Algnameabbr{} outperforms other methods on most of the datasets, with the exception of the X dataset where it attains the second place. 

Nevertheless, ground-truth labels are not accessible in real-world datasets, such as ogbn-arxiv. 
Therefore, to assess the explanations in a more general way, fidelity scores are employed. Recent works on GNN explanations have begun to prioritize fidelity over ground-truth due to this reasons~\cite {lin2022orphicx,zhang2022gstarx}. As such, we evaluate explanation faithfulness using fidelity v.s. sparsity measures in the main paper, which is more general and challenging.

\begin{table}[h]
\centering
\begin{tabular}{cccccc}
\toprule[1pt]
           & BA-Shapes & BA-Community & Tree-Cycles & BA-2Motifs  \\\hline
GNNExplainer   & 0.448  &   0.441     & 0.612    &      0.487   \\
PGExplainer  & \textbf{0.517} &     0.443     &  0.626      & 0.268          \\
GraphSVX        & 0.447 &   0.441      &   0.611     &    0.492       \\ 
LARA        & 0.466  &    \textbf{0.475}     &   \textbf{0.641}    & \textbf{0.913}           \\
\bottomrule[1pt]
\end{tabular}
\caption{AUC of LAVA compared to other baselines on BA-Shapes, BA-Community, Tree-Cycles, and BA-2Motifs datasets. The best result is in bold.}
\label{tab:auc}
\end{table}

\section{More Qualitative Results}
\label{supp:quali}

\begin{figure*}[h!]
\begin{center}
        \begin{subfigure}[b]{0.24\textwidth}
        \centering
                \includegraphics[width=.5\linewidth]{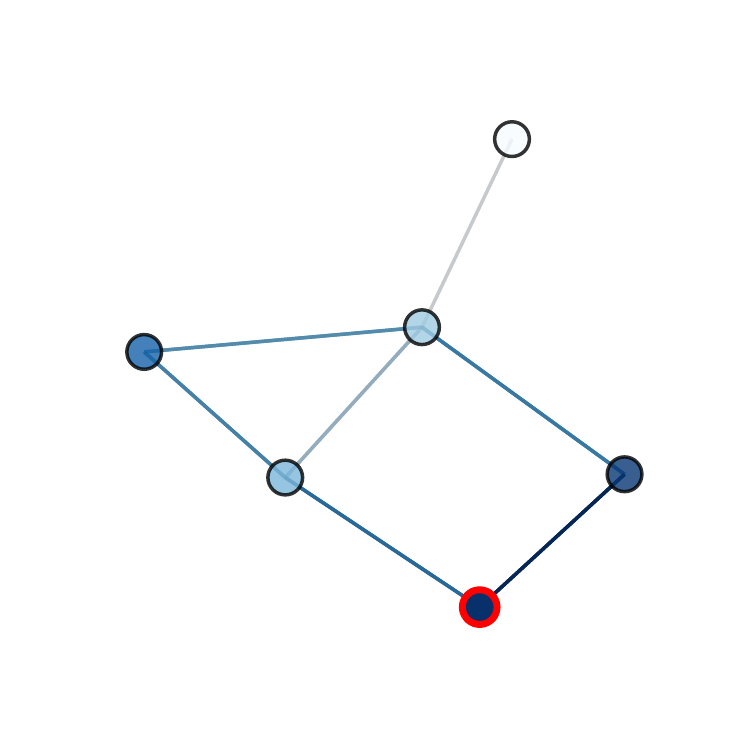}
                \caption{BA-Shapes}
                \label{subfig:bashapes0}
        \end{subfigure}%
        \hfill
        \begin{subfigure}[b]{0.24\textwidth}
        \centering
                \includegraphics[width=.5\linewidth]{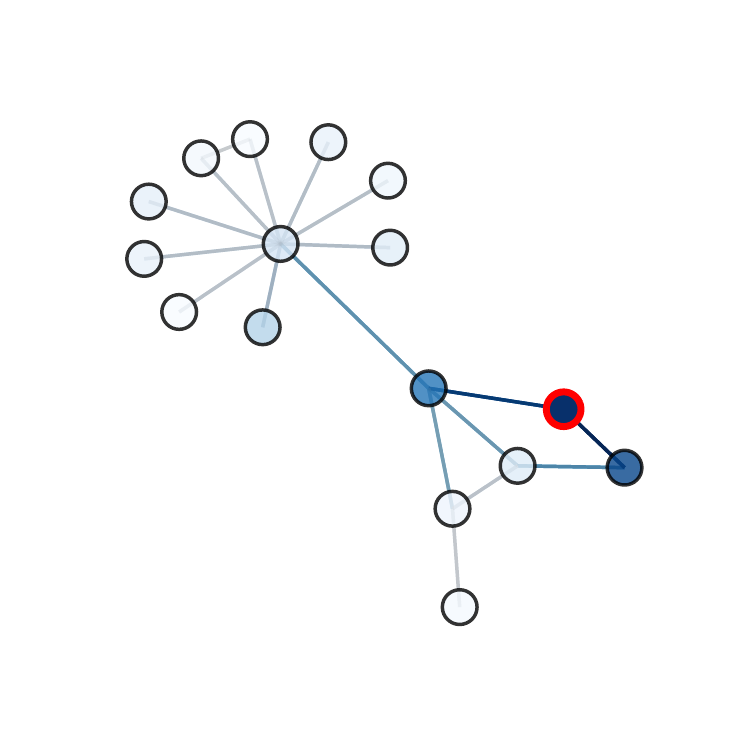}
                \caption{BA-Community}
                \label{subfig:bacomm0}
        \end{subfigure}%
        \hfill
        \begin{subfigure}[b]{0.24\textwidth}
        \centering
                \includegraphics[width=.5\linewidth]{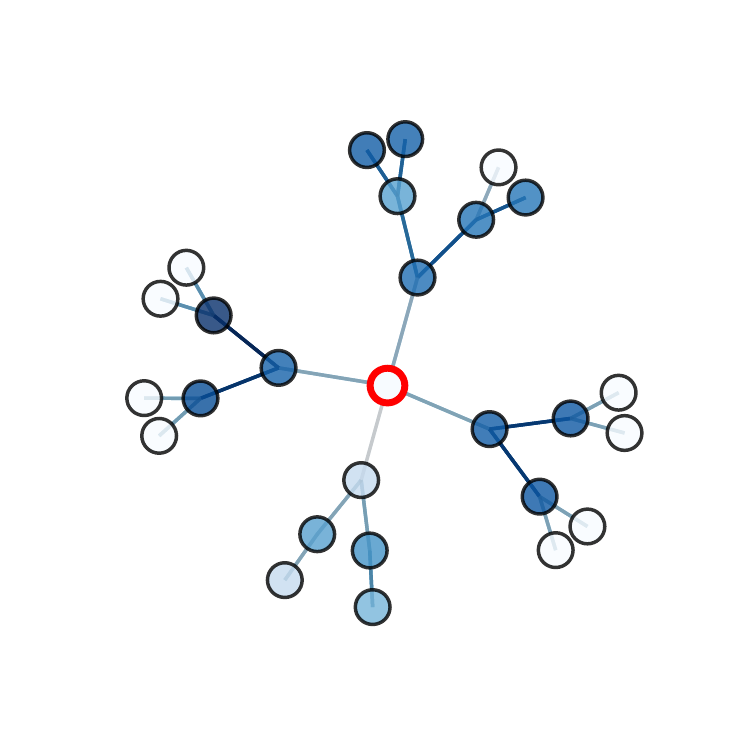}
                \caption{Tree-Cycles}
                \label{subfig:tree0}
        \end{subfigure}%
        \hfill
        \begin{subfigure}[b]{0.24\textwidth}
        \centering
                \includegraphics[width=.5\linewidth]{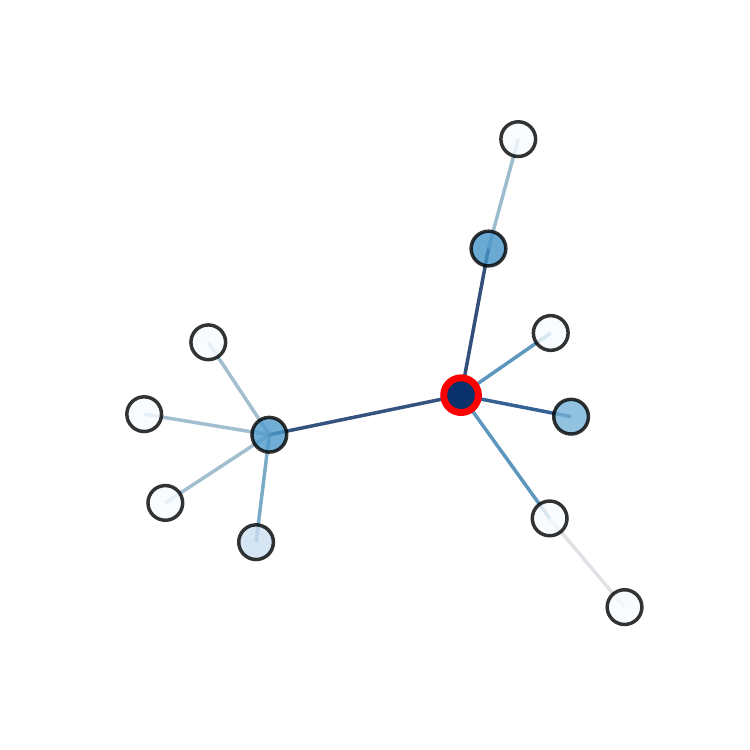}
                \caption{ogbn-arxiv}
                \label{subfig:arxiv0}
        \end{subfigure}
        \\
        \begin{subfigure}[b]{0.24\textwidth}
        \centering
                \includegraphics[width=.5\linewidth]{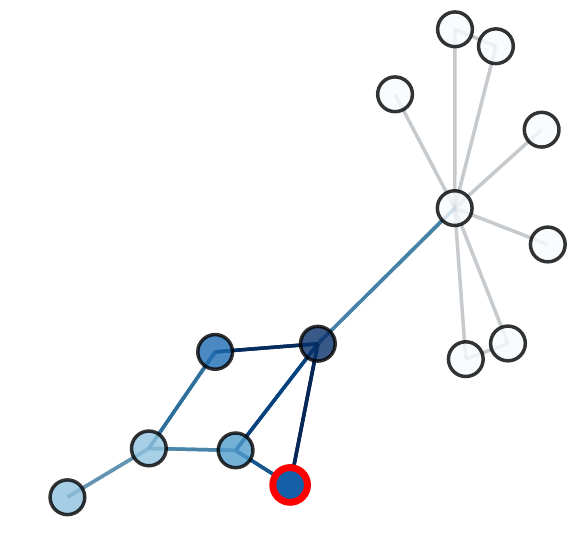}
                \caption{BA-Shapes}
                \label{subfig:bashapes1}
        \end{subfigure}
        \hfill
        \begin{subfigure}[b]{0.24\textwidth}
        \centering
                \includegraphics[width=.5\linewidth]{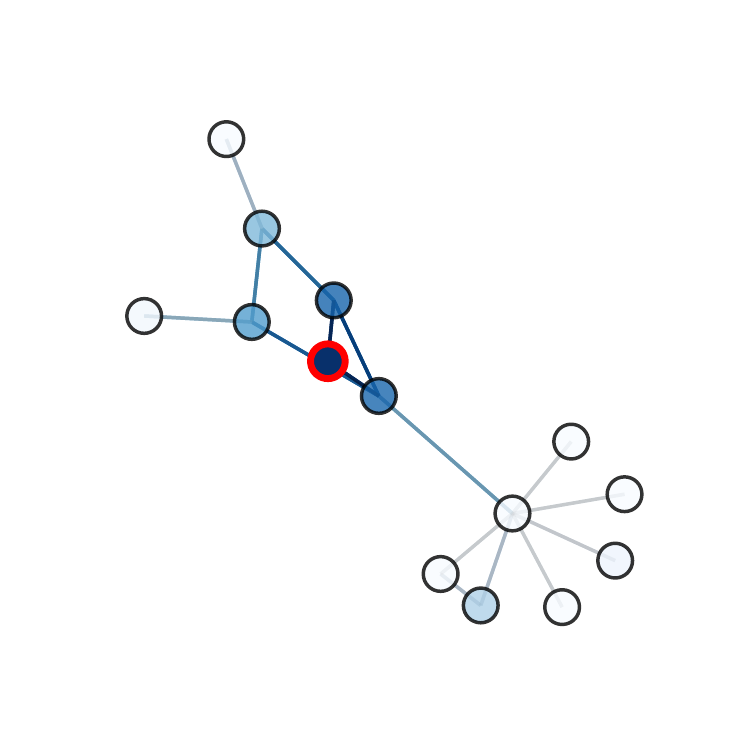}
                \caption{BA-Community}
                \label{subfig:bacomm1}
        \end{subfigure}%
        \hfill
        \begin{subfigure}[b]{0.24\textwidth}
        \centering
                \includegraphics[width=.5\linewidth]{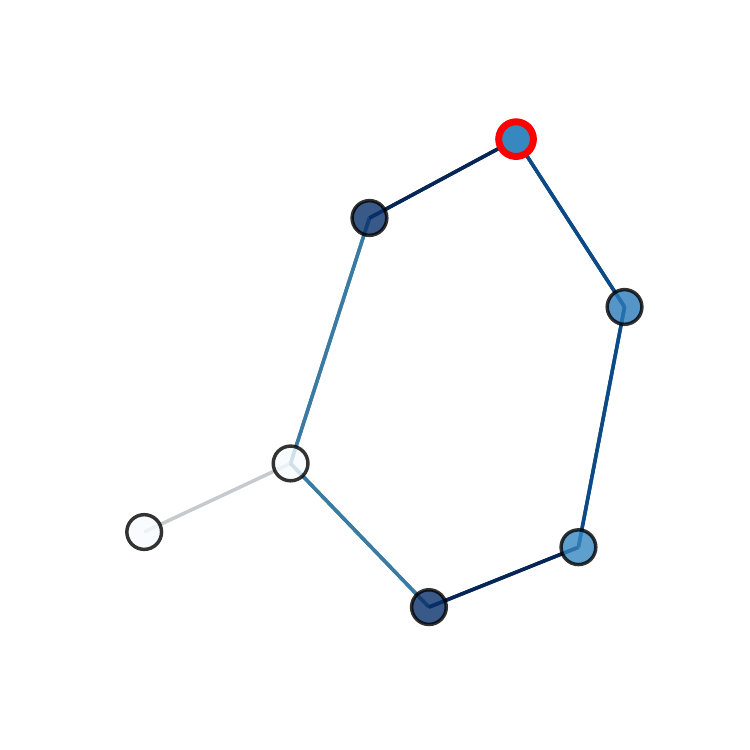}
                \caption{Tree-Cycles}
                \label{subfig:tree1}
        \end{subfigure}%
        \hfill
        \begin{subfigure}[b]{0.24\textwidth}
        \centering
                \includegraphics[width=.5\linewidth]{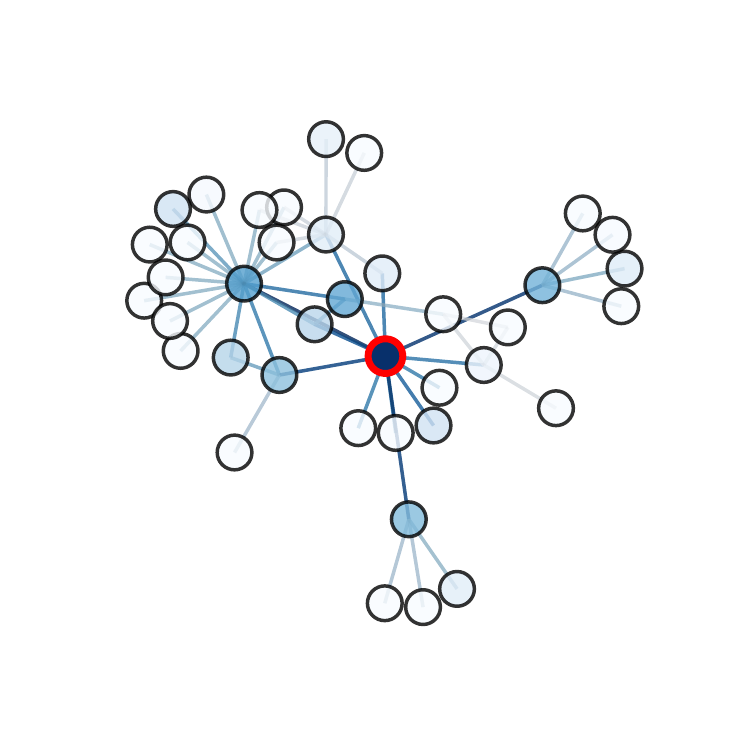}
                \caption{ogbn-arxiv}
                \label{subfig:arxiv1}
        \end{subfigure}%
        \caption{More qualitative results on node classification tasks. The color of the nodes indicates the importance score (explanations). Darker color refers to a higher score. Two examples are shown for each dataset. The target node (to be explained) is highlighted in red color. 
        }
        \label{app-fig:quali node cls}
\end{center}
\end{figure*}

\begin{figure*}[h!]
\begin{center}
        \begin{subfigure}[b]{0.23\textwidth}
        \centering
                \includegraphics[width=.6\linewidth]{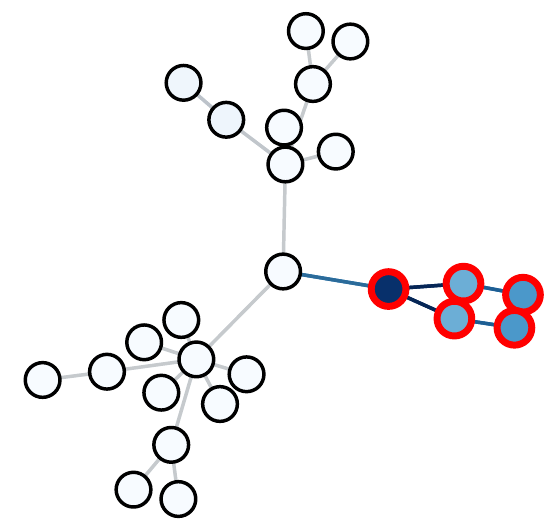}
                \caption{BA-2Motifs, Pentagon}
                \label{subfig:ba2-0-0}
        \end{subfigure}%
        \hfill
        \begin{subfigure}[b]{0.23\textwidth}
        \centering
                \includegraphics[width=.6\linewidth]{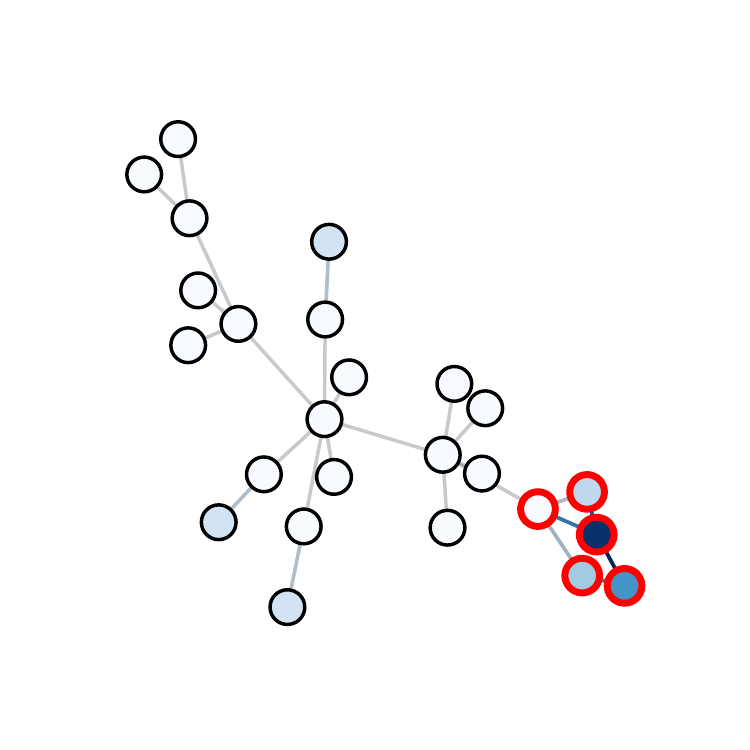}
                \caption{BA-2Motifs, House}
                \label{subfig:ba2-1-0}
        \end{subfigure}%
        \hfill
        \begin{subfigure}[b]{0.23\textwidth}
        \centering
                \includegraphics[width=.6\linewidth]{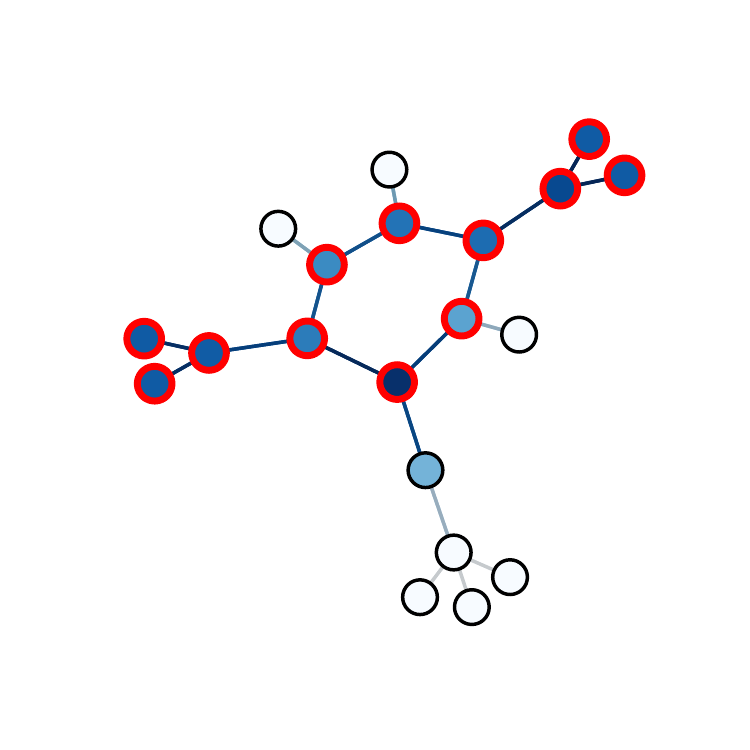}
                \caption{MUTAG, Mutagen}
                \label{subfig:mutag-0-0}
        \end{subfigure}%
        \hfill
        \begin{subfigure}[b]{0.23\textwidth}
        \centering
                \includegraphics[width=.6\linewidth]{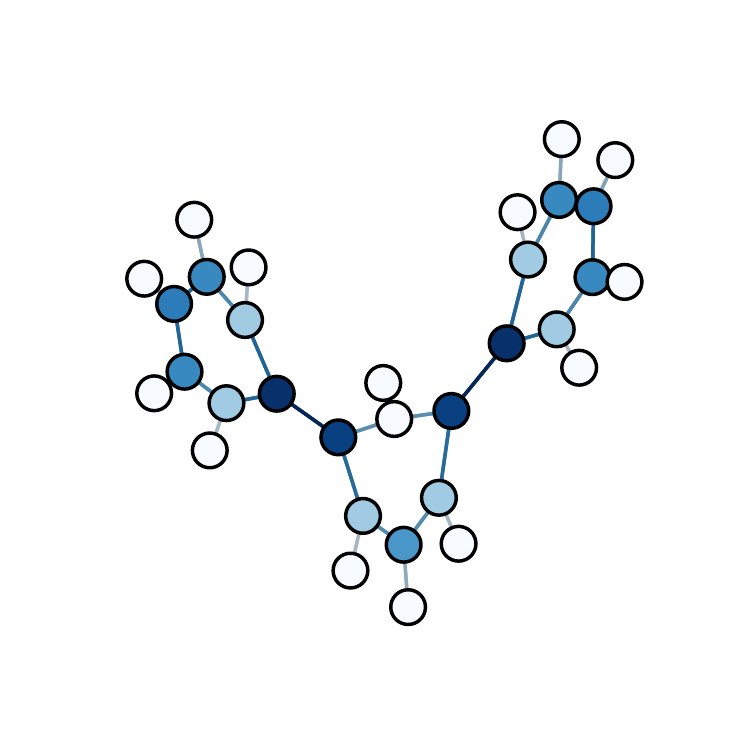}
                \caption{MUTAG, Non-Mutagen}
                \label{subfig:mutag-1-0}
        \end{subfigure}
        \\
        \begin{subfigure}[b]{0.23\textwidth}
        \centering
                \includegraphics[width=.6\linewidth]{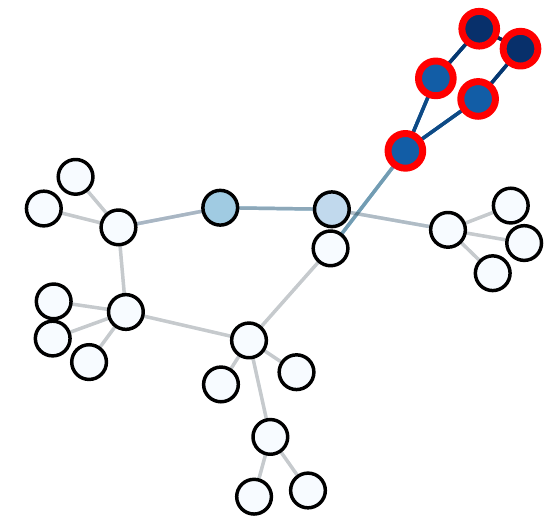}
                \caption{BA-2Motifs, Pentagon}
                \label{subfig:ba2-0-1}
        \end{subfigure}
        \hfill
        \begin{subfigure}[b]{0.23\textwidth}
        \centering
                \includegraphics[width=.6\linewidth]{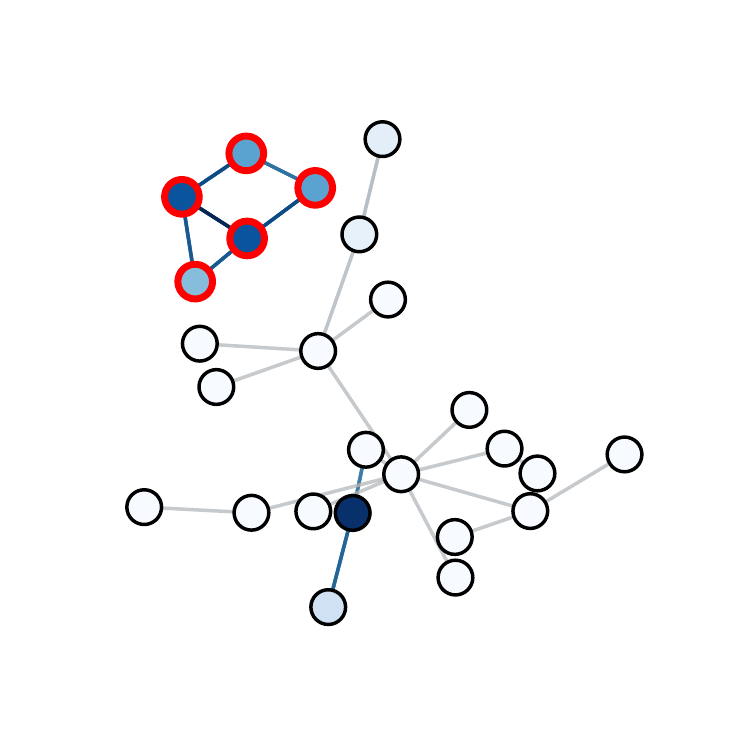}
                \caption{BA-2Motifs, House}
                \label{subfig:ba2-1-1}
        \end{subfigure}%
        \hfill
        \begin{subfigure}[b]{0.23\textwidth}
        \centering
                \includegraphics[width=.6\linewidth]{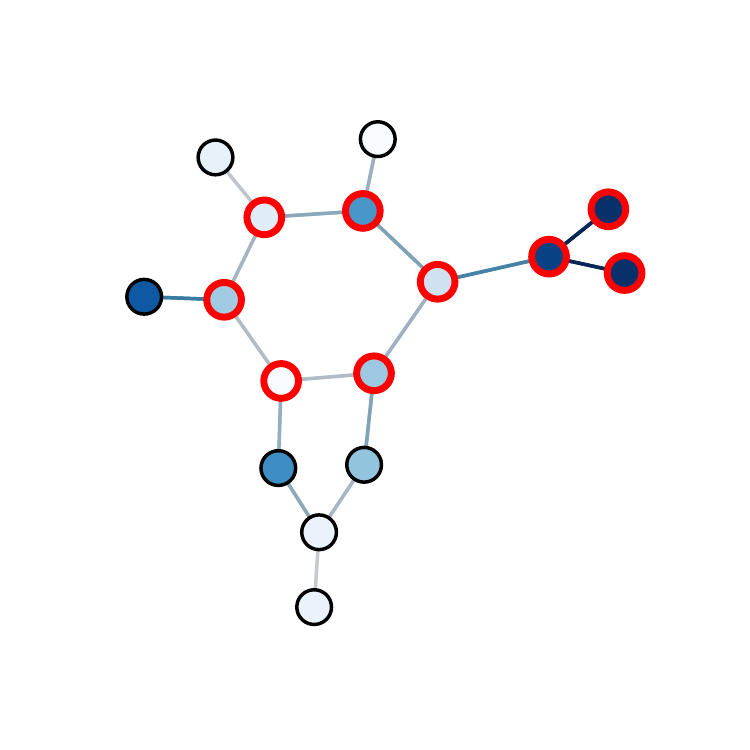}
                \caption{MUTAG, Mutagen}
                \label{subfig:mutag-0-1}
        \end{subfigure}%
        \hfill        
        \begin{subfigure}[b]{0.23\textwidth}
        \centering
                \includegraphics[width=.6\linewidth]{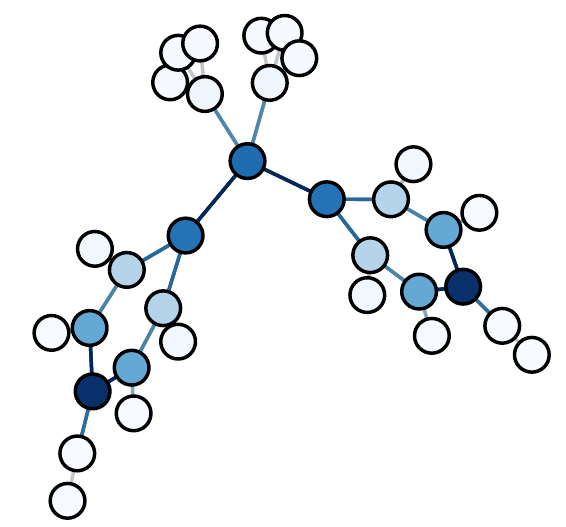}
                \caption{MUTAG, Non-mutagen}
                \label{subfig:mutag-1-1}
        \end{subfigure}%
        \caption{More qualitative results on graph classification tasks. The color of the nodes indicates the importance score (explanations). Darker color refers to a higher score. The dataset name and class of each instance are given. 
        The ground truth motif is highlighted in red. On MUTAG, NH$_2$/NO$_2$ together with the carbon ring is the ground truth for the class mutagen. while the class Non-mutagen is without this pattern.
        }
        \label{app-fig:quali graph cls}
\end{center}
\end{figure*}

We present more examples of the explanations given by \Algnameabbr{} on node classification and graph classification tasks in \Cref{app-fig:quali node cls} and \Cref{app-fig:quali graph cls}, respectively.

\section{Releted Work}

\subsection{GNN Explanations}
The model explanation is one of the emerging technologies in recent years due to the wide application of machine learning models, where a handful of explanation methods are proposed specifically for GNNs. 
Popular existing works can be divided into gradient-based, surrogate-based, decomposition-based and perturbation-based methods~\cite{yuan2022explainability}. Gradient-based explanations follow a similar procedure as on image and text data. For instance, \citet{pope2019explainability} propose CAM and GradCAM for graphs, which make use of the final fully-connected layer in the graph model to produce the feature map for nodes. 
Surrogate-based methods explain the original model via approximating it using an interpretable model~\cite{vu2020pgm,huang2022graphlime,zhang2021relex}. For example, the feature weights given by the algorithm Hilbert-Schmidt Independence Criterion Lasso \cite{yamada2014high} are used as explanations for the original GNN in \cite{huang2022graphlime}. Decomposition-based methods back-propagate the prediction score layer by layer into the input space following the designed decomposition
rules~\cite{feng2021degree,pope2019explainability,schnake2021higher}, while perturbation-based explanations assign importance scores to nodes or edges by observing output changes when masking different input features~\cite{ying2019gnnexplainer,luo2020parameterized,funke2020hard,yuan2021explainability, duval2021graphsvx, zhang2022gstarx, lin2021generative}. 
Most of the perturbation-based methods~\cite {ying2019gnnexplainer,luo2020parameterized,lin2022orphicx,lin2021generative} learn masks of edges by minimizing the mutual information-relevant loss between the predictions with and without masking, while \cite{yuan2021explainability, duval2021graphsvx, zhang2022gstarx} provide explanatory nodes via estimating importance scores of nodes. Among these works, only a few works such as \cite{luo2020parameterized,lin2022orphicx,lin2021generative} focus on amortized explainers that can provide explanations on a test set in an inductive way after training. Different from previous amortized explainers for GNNs, we propose a novel algorithm utilizing removal-based attribution as guidance to train an explainer. In this work, we demonstrate the advantage of our method in both explanation faithfulness and efficiency compared to previous methods.


\subsection{Amortized Explanations}
In amortized explanation methods, a learned explainer generates explanations for a batch of instances via a forward pass~\cite{covert2021explaining}. \cite{chuang2022cortx,jethani2021fastshap,chen2018learning,jethani2021have} showcase real-time explanations on image and tabular data. To be concrete, \citet{jethani2021fastshap} 
propose an algorithm FastSHAP, where an objective function is designed for an explainer to capture Shapley value distribution, achieving real-time explanations. CoRTX~\cite{chuang2022cortx} trains an explainer using positive and negative samples by minimizing the contrastive loss. Moreover, very few labels (ground truth explanations) are used during training to boost the interpretation performance.
\citet{chen2018learning} realize an explainer returning a distribution over selected features given the input instance. The explainer is trained by maximizing the variational approximation of mutual information between selected features and the response variable from the target model. 
\citet{jethani2021have} propose to train a global selector and predictor jointly, where the selector (REAL-X) identifies the minimal subset of features to maximize the fidelity evaluated by a predictor model (EVAL-X).
Compared to previous amortized explanations on regular data, our explainer has no constraints on input shape (unlike image or tabular data) and adapts to irregular data (graph data). Furthermore, our proposed objective function for learning removal-based attribution is designed to enable high-fidelity explanation generation. We validate its advantage from both theoretical and experimental perspectives in this work.

\section{Limitation and Potential Negative Social Impact}

In this work, we deploy an amortized explainer to learn the removal-based attribution such that the GNN explanation can be generated fast in practical application scenarios.
We remark that implementing and training the amortized explainer would have carbon emissions to the environment. Nevertheless, we wish to minimize the need for repetitive experiments with our theoretical analysis.
